\documentclass[runningheads]{llncs}

\usepackage{eccv}

\usepackage{eccvabbrv}

\usepackage{graphicx}
\usepackage{booktabs}

\usepackage[accsupp]{axessibility}  

\usepackage{multirow}
\usepackage{lipsum}

\usepackage{bm,cuted}
\usepackage[ruled,vlined,linesnumbered]{algorithm2e}
\usepackage{graphicx}
\usepackage{tabularx} 
\usepackage[safe]{tipa}
\usepackage{multirow}

\usepackage{pifont}
\makeatletter
\newcommand\HUGE{\@setfontsize\Huge{20}{30}}
\makeatother

\def\eqref#1{equation~\ref{#1}}

\def\1{\bm{1}}

\DeclareMathAlphabet{\mathsfit}{\encodingdefault}{\sfdefault}{m}{sl}
\SetMathAlphabet{\mathsfit}{bold}{\encodingdefault}{\sfdefault}{bx}{n}

\usepackage{soul}
\usepackage{enumitem}
\setlist[itemize]{topsep=3pt}

\usepackage[bookmarks=false, linkcolor=blue, urlcolor=blue, citecolor=blue]{hyperref} 
\hypersetup{
    colorlinks=true,
    linkcolor=red,
    filecolor=magenta,      
    urlcolor=blue,
    pdfstartview={FitH},
    citecolor =blue
    }
    
\usepackage{orcidlink}

\begin{document}

\title{Scalable Group Choreography via Variational Phase Manifold Learning} 


\author{Nhat Le\inst{1}\orcidlink{0009-0007-1122-4981} \and
Khoa Do\inst{2,3}\orcidlink{0009-0008-3819-1312} \and
Xuan Bui\inst{2,3} \and
Tuong Do\inst{1,4}\orcidlink{0000-0002-3290-3787} \and
Erman Tjiputra\inst{1}\orcidlink{0009-0003-6909-4623} \and
Quang D.Tran\inst{1}\orcidlink{0000-0001-5839-5875} \and
Anh Nguyen\inst{4}\orcidlink{0000-0002-1449-211X}}

\authorrunning{N.~Le et al.}

\institute{AIOZ, Singapore 
\and
University of Science, Ho Chi Minh City, Vietnam
\and
Vietnam National University, Ho Chi Minh City, Vietnam
\and 
University of Liverpool, Liverpool L69 3BX, United Kingdom
}

\maketitle

\begin{abstract}
Generating group dance motion from the music is a challenging task with several industrial applications. Although several methods have been proposed to tackle this problem, most of them prioritize optimizing the fidelity in dancing movement, constrained by predetermined dancer counts in datasets. This limitation impedes adaptability to real-world applications. Our study addresses the scalability problem in group choreography while preserving naturalness and synchronization. In particular, we propose a phase-based variational generative model for group dance generation on learning a generative manifold. Our method achieves high-fidelity group dance motion and enables the generation with an unlimited number of dancers while consuming only a minimal and constant amount of memory. The intensive experiments on two public datasets show that our proposed method outperforms recent state-of-the-art approaches by a large margin and is scalable to a great number of dancers beyond the training data.
\vspace{-2ex}
\begin{figure}[ht]
\begin{center}
  \centering
\resizebox{\linewidth}{!}{
\setlength{\tabcolsep}{2pt}
\begin{tabular}{cccc}
\shortstack{\includegraphics[width=0.32\linewidth]{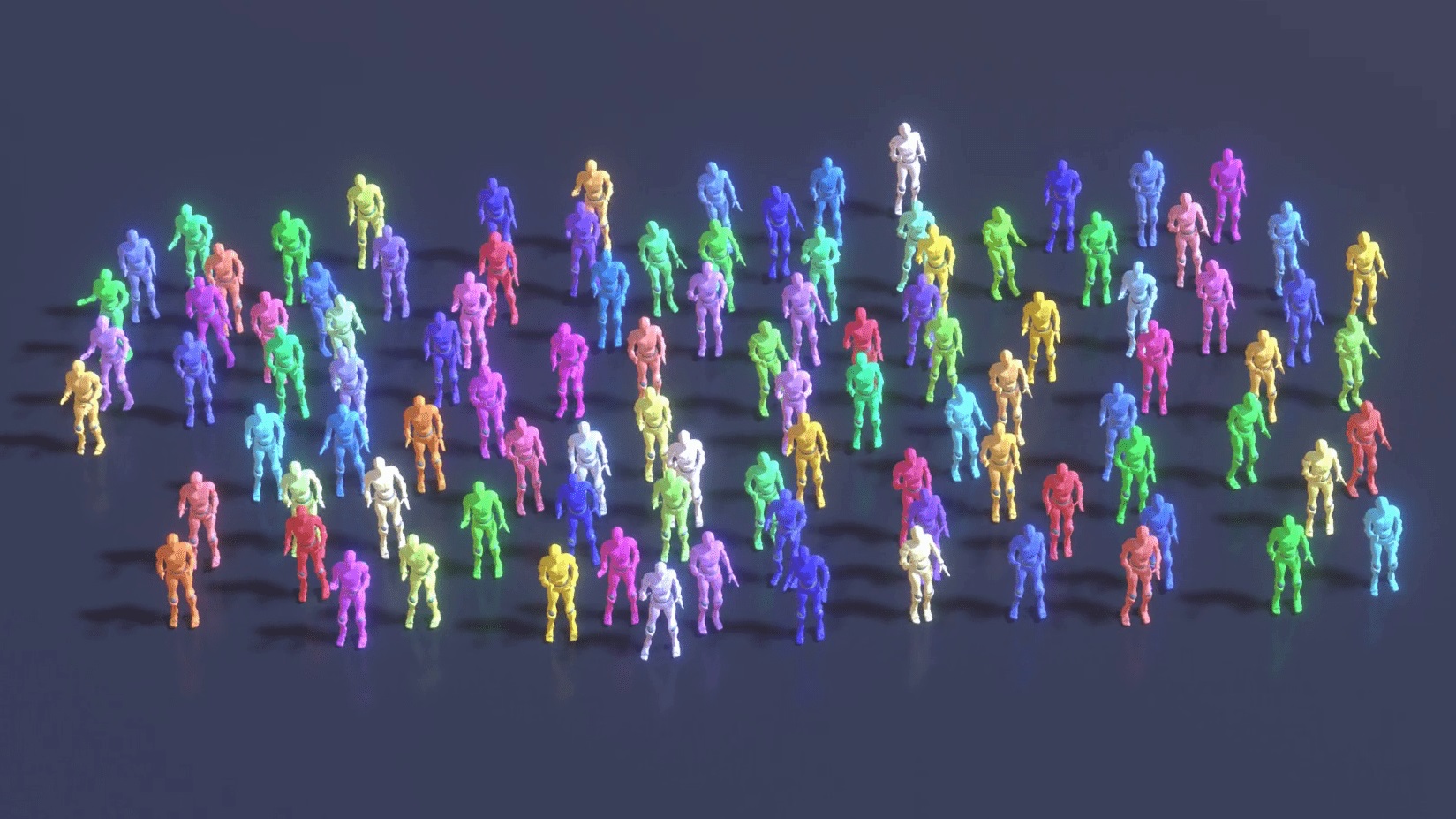}}&
\shortstack{\includegraphics[width=0.32\linewidth]{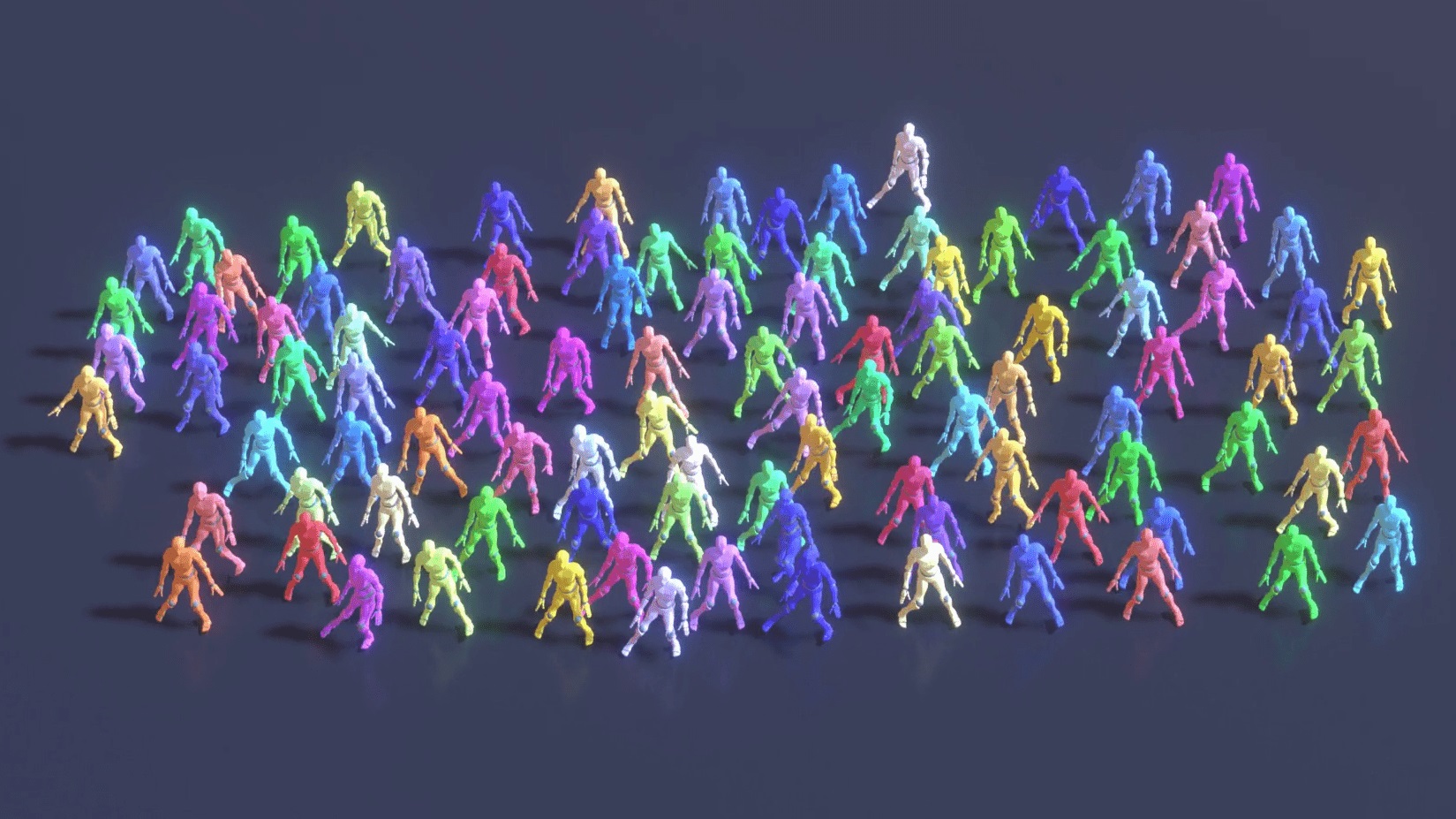}}&
\shortstack{\includegraphics[width=0.32\linewidth]{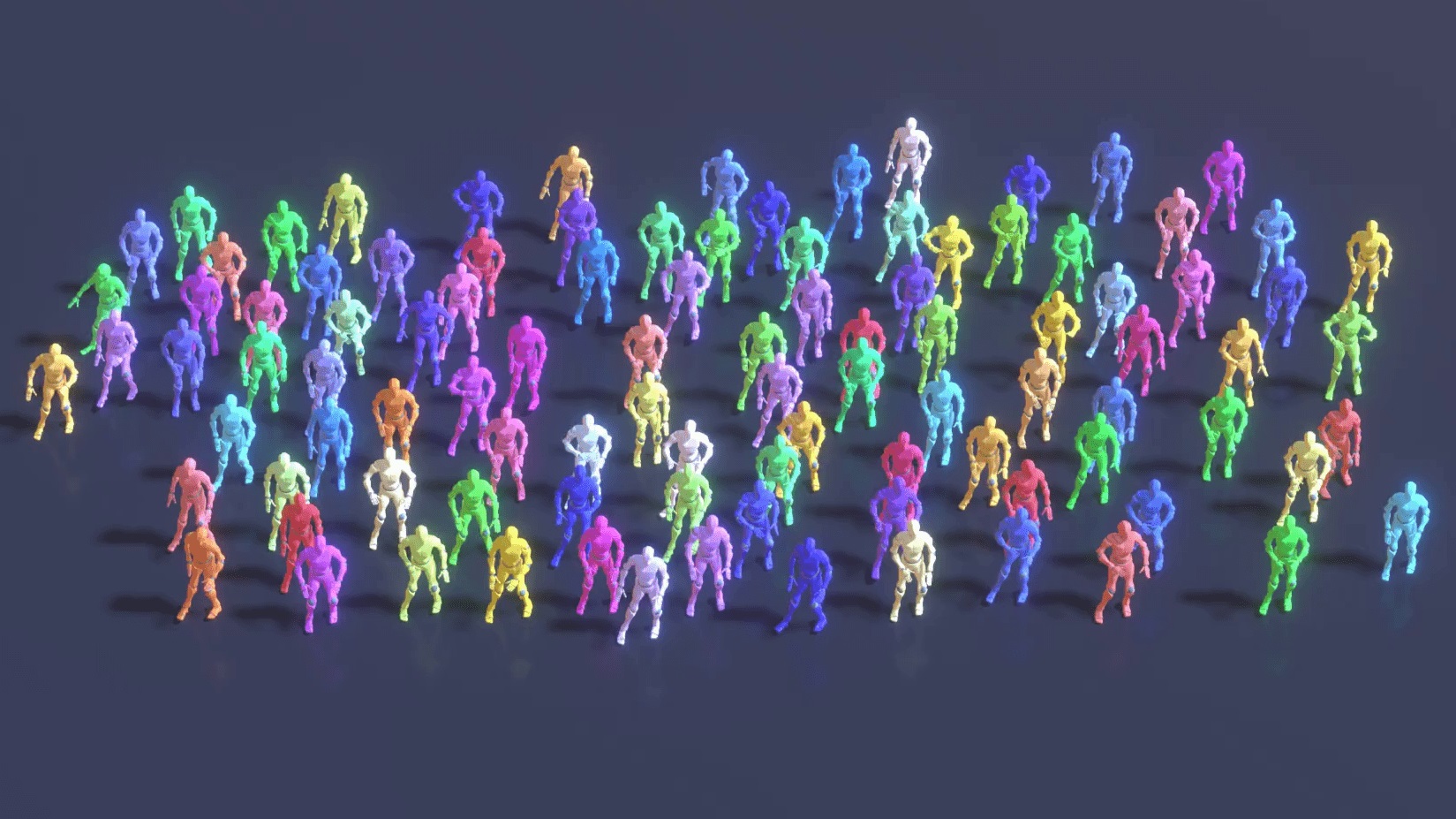}}&

\end{tabular}
}
\vspace{-2ex}
    \caption{We present a new group dance generation method that can generate a large number of dancers within a fixed resource consumption. The illustration shows a generated group dance sample with $100$ dancers.}
     \label{fig:teaser}
\end{center}%
\end{figure}

\end{abstract}
\vspace{-16ex}

\section{Introduction}
\vspace{-1ex}
The widespread availability of digital social media platforms has led to a significant increase in the popularity of creating dance videos. This heightened interest has resulted in the daily production and viewing of millions of dance videos on various online platforms~\cite{FINK2021351, Visualization-folk-dance}. Recently, researchers from the computer vision and computer graphics community have focused on developing methods for generating authentic dance movements in response to music~\cite{bisig2022generative, zhang2023mining}. These advancements have broad implications and find applications in diverse areas, including animation~\cite{li2021AIST++, li2024controlling}, virtual idols~\cite{perez2021_transflower}, virtual metaverses~\cite{survey-dancing-deep-metaverse}, and dance education~\cite{physic-perform, soga2005automatic, shi2021application}. Such techniques empower artists, animators, and educators by providing powerful tools to enhance their creative pursuits and improve the overall dance experience for performers and audiences alike.

Recently, significant strides have been made in creating dance motions for solo dance performers~\cite{li2021AIST++,ferreira2021_learn2dance_gcn,perez2021_transflower,siyao2022_bailando,kim2022brandnew_dance,Dance_Revolution,tseng2022edge_diffusion}. However, generating synchronized group dance movements that are both lifelike and in harmony with music remains a complex challenge~\cite{yalta2019_weaklyrnn,chen2021_choreomaster, huang2022genre}. The introduction of AIOZ-GDance~\cite{le2023music} stands as the extensive dataset to aid in generating group choreography. Different aspects of group dance motion such as the study of consistency and diversity among the movements of dancers are explored in~\cite{le2023controllableGCD}. However, despite recent advances in the field, existing methods can either only generate dances for a limited number of dancers given the input audio~\cite{yao2023dance,wang2022groupdancer,le2023music}, or would consume an expensively large amount of memory due to their design nature~\cite{le2023music,le2023controllableGCD}. Typically, these methods are usually based on specialized collaborative mechanisms such as cross-entity attention~\cite{le2023music,liang2023intergen}, group global attention~\cite{le2023controllableGCD}, or residual communication block~\cite{shafir2023prior_mdm_two_person} achieve coherent multi-person synthesis. In other words, their innate architectural design requires the network to process and synthesize all motions simultaneously, leading to excessively growing computational/memory costs with the increasing number of inputs, and hence extremely difficult to scale up. 
                                                                        

Despite the comprehensive exploration of group dance motion generation, the predominant emphasis remains on optimizing the execution of the generated movements~\cite{lin2024motion, gartner2023transformer, kolotouros2024dreamhuman, yu2023monohuman, he2022nemf, karunratanakul2023guided}. These existing methodologies function within a confined scope, often tethered to a predetermined maximum number of dancers depicted in group choreography videos outlined by the predefined dataset~\cite{le2023controllableGCD}. This inherent limitation poses a potential constraint on the adaptability of trained models when transitioning to real-world applications~\cite{li2021audio2gestures}. In our pursuit, we center our efforts on addressing the critical aspect of \textit{scalability in group dance}. 
Our primary objective revolves around not only expanding the number of dancers but also ensuring the preservation of the innate naturalness in the dance motions and the seamless synchronization between dancers throughout the entirety of their performance.



In the field of character motion control and synthesis, existing approaches can be broadly divided into two categories: deterministic~\cite{martinez2017human,ghosh2021synthesis,Dance_Revolution,li2021AIST++,le2023music,wang2022groupdancer} and probabilistic~\cite{he2022nemf,petrovich2022temos_t2m_vae,perez2021_transflower,tseng2022edge_diffusion,le2023controllableGCD}. The former tries to learn direct and deterministic mapping from the input conditioning signal (such as text, audio, or user controls) to the desired output motion. However, these methods usually regress towards the mean pose as they are typically trained by minimizing a regression objective under different outputs for similar inputs, resulting in freezing or drifting motion. On the other hand, probabilistic methods attempt to capture the distribution over possible motions for a given condition, allowing diverse and high-quality generation. In particular, Variational Autoencoder (VAE)~\cite{petrovich2021action,petrovich2022temos_t2m_vae,li2021audio2gestures,guo2022text2motion_vae}, or Diffusion Models~\cite{tevet2022mdm,tseng2022edge_diffusion,zhou2023ude_diffusion_motion,alexanderson2023diffusion, le2023controllableGCD} have been widely used. Although diffusion models have recently shown strong potential in generating high-quality motions~\cite{tevet2022mdm}, they generally operate on high-dimensional space (same as the original data)~\cite{tevet2022mdm,tseng2022edge_diffusion,le2023controllableGCD}, hence rendering them extremely challenging to use for scalable group generation. Meanwhile, standard VAE models often rely on a Gaussian latent space of fixed dimension for sampling a single latent vector to generate motions, making it difficult to adapt to a wide range of architectures. Besides, the fixed-size single latent space might also not be sufficient to represent the temporal dynamics of the whole motion sequence~\cite{cervantes2022implicit_motion,petrovich2021action}.

Recent studies have demonstrated the significant advantages of learning motion features in the frequency domain. Specifically, \cite{holden2017phase_fnn,starke2019neural_state_machine,starke2020local_motion_phase,starke2022deepphase} have found that different motion skills can be effectively represented by several phase variables that can faithfully capture the spatial-temporal alignment of a wide range of movements. This can be used to combine with common deep motion synthesis networks to enhance its capability, facilitating natural, and stable motion generations by enforcing a unidirectional motion transition (i.e., no backward direction in time) to prevent the motion from being stuck temporally~\cite{starke2022deepphase}. Unlike those prior works, where the phases are utilized as auxiliary signals to assist in the motion synthesis process~\cite{starke2022deepphase,yang2023qpgesture,shi2023phasemp}, we propose to learn to generate the phases conditioned on the input audio signal in an end-to-end manner, eliminating the need of training extra components. Our generative phase manifold can be used to generate new motions efficiently, from which the latent motion curves are sampled in frequency domain and decoded into the original motion space by a learned decoder.

In this paper, our goal is to develop a scalable technique for group dance generation, a phase-based variational generative model for scalable group dance generation, namely Phase-conditioned Dance VAE (PDVAE). To our knowledge, PDVAE is the first method to represent the variational latent space using phase parameters in the frequency domain of the motion curves.  Our method goes beyond the conventional VAE approach that typically relies on a single latent vector drawn from a Gaussian distribution, which is unable to adequately represent the temporal information of the motion sequence (e.g., the time dimension is squeezed out). 
Figure~\ref{fig:teaser} illustrates an example of a dance motion sequence with 100 dancers generated by our model. In contrast to existing methods, our model can generate crowd-dance animations for an unlimited number of dancers without increasing computational burden (i.e., memory consumption remains constant), while still maintaining the performance and fidelity of the generated choreographies. 
To summarize, our key contributions are as follows:
\begin{itemize}
    \item We introduce PDVAE, a phase-based variational generative model for scalable group dance generation. The method focuses on generating large-scale group dance under limited resources.
    
    \item To effectively learn the manifold that is group-consistent (i.e., dancers within a group lie upon the same manifold), 
    we propose a group consistency loss that enforces the networks to encode the latent phase manifold to be identical for the same group given the input music.
    
    \item Extensive experiments along with thorough user study evaluations demonstrate the state-of-the-art performance of our model while achieving effective scalability.
\end{itemize}

\section{Related Work}
\label{Sec:relatedwork}
\subsection{Music-driven Choreography} 
Crafting natural human choreography derived from music presents a multifaceted challenge~\cite{joshi2021extensive, yang2023keyframe,qi2023diffdance}. Certain methodologies integrate constraints based on music-motion similarity matching to ensure coherence between the generated motion and the music~\cite{motion_graph2, motion_graph3, motion_graph4, motion_graph5, motion_graph6, motion_graph7}. However, many of these approaches rely considerably on heuristic algorithms, stitching together pre-existing dance segments sourced from a limited music-dance database~\cite{motion_graph7, li2023finedance}. While successful in generating extended and realistic dance sequences, these methods face limitations when endeavoring to create entirely novel dance fragments~\cite{ofli2011learn2dance, yao2023dance}.

In the recent period, advancements have been evident in the domain of converting music into dance movements through various techniques such as Convolutional Networks (CNN)\cite{chan2019everybody, zhuang2020_music2dance, sun2020_deepdance, ye2020_choreonet, ahn2020_autoregressive, yin2023dance}, Recurrent Networks (RNN)\cite{tang2018_dancemelody, sun2020_deepdance, alemi2017_groovenet, Dance_Revolution, yalta2019_weaklyrnn}, Graph Neural Networks (GNN)\cite{ferreira2021_learn2dance_gcn, ren2020_ssl_gcn, au2022choreograph, zhou2022spatio_temporal_gcn}, Generative Adversarial Networks (GAN)\cite{sun2020_deepdance, lee2019_dancing2music}, or Transformer models~\cite{siyao2022_bailando, li2021AIST++, li2022_phantomdance, perez2021_transflower, kim2022brandnew_dance, li2022danceformer}. These methodologies typically depend on various inputs, including the present music and a concise history of previous dance movements, to forecast forthcoming sequences of human poses. 
This innovation introduces a music-text feature fusion module, engineered to amalgamate inputs into a motion decoder, thereby enabling the creation of dance sequences conditioned upon both musical and textual instructions~\cite{gong2023tm2d}.

While these techniques show promise in generating authentic and lifelike dance motions, they often struggle to synchronize movements seamlessly across multiple dancers~\cite{le2023music}. Specifically, achieving coordination and harmony among dancers necessitates consideration of spatial and temporal relationships, their interactions, and the overall choreographic structure~\cite{AISTDanceDB}. Consequently, further advancements in this domain are being pursued to tackle these complexities~\cite{li2021audio2gestures, hong2022avatarclip, zhu2022quantized, perez2021_transflower}. For instance, Perez \etal~\cite{perez2021_transflower} integrate a multimodal transformer encoder with a normalizing-flow-based decoder to estimate a probability distribution encompassing potential subsequent poses.
Additionally, Feng \etal~\cite{feng2023robust} enable long-term generation by imposing a motion repeat constraint to forecast future frames while considering historical motions.
A recent study by Le \etal~\cite{le2023controllableGCD} investigates consistency and diversity factors between the generated motions of two or more dancers within a timeframe. However, these approaches are limited to a predefined number of dancers, heavily constrained by the maximum number of dancers in the dataset-provided videos.

\subsection{Motion Manifold Learning}
Motion manifold learning has attracted considerable attention in computer vision and artificial intelligence, with the primary goal of comprehending the fundamental structures inherent in human movement and dynamics~\cite{holden2015learning, mo2023continuous, ghosh2021synthesis, tiwari2022pose}. Its distinctive ability to generate human movement patterns presents numerous opportunities to comprehend intrinsic motion dynamics, manage nonlinear relationships in motion data, and acquire contextual and hierarchical representations~\cite{jiang2023drop, yang2023qpgesture, raab2023modi}. Consequently, various methodologies have emerged, each contributing distinct perspectives and techniques to advance the comprehension and synthesis of human motion~\cite{jiang2023motiondiffuser,raab2023modi, shi2023phasemp, lannan2022human}.

For instance, Holden \etal~\cite{holden2016deep} generate character movements by mapping high-level parameters to the human motion manifold, allowing diverse movements without manual preprocessing, and enabling post-generation editing for natural, smooth motion sequences. MotionCLIP~\cite{tevet2022motionclip}  introduces a 3D human motion auto-encoder aligned with the Contrastive Language-Image Pre-training (CLIP) model's space, enabling semantic text-based motion generation, disentangled editing, and abstract language specification, leveraging CLIP's rich semantic knowledge within the motion manifold. Recently, Sun \etal~\cite{sun2022you}  utilize a VQ-VAE to acquire a low-dimensional manifold, effectively cleansing the motion sequences.
Regarding the scalability of group dance generation, the concept of the motion manifold emerges as a potential solution for addressing the restricted number of dancers in the dataset. This direction enables the learning of a distribution that facilitates the extraction of dance motions, allowing for the presentation of cohesive group dance sequences.

\section{Variational Phase Manifold Learning for Scalable Group Choreography}

\subsection{Task Definition}
Given an input music sequence $\mathbf{a} = \{a_1, a_t, ...,a_T\}$ with $t = \{1,..., T\}$ indicates the index of the music frames, our goal is to generate the group motion sequences of $N$ arbitrary dancers: $\mathbf{x} = \{x^1_1,..., x^1_T; ...;x^N_1,...,x^N_T\}$ where $x^n_t$ is the pose of $n$-th dancer at frame $t$. 
We follow~\cite{ling2020motion_vae,he2022nemf} and represent dance as a sequence of poses in the 24-joint of the SMPL body model~\cite{SMPL:2015}, using the 6D continuous rotation~\cite{zhou2019rotation_6d} for every joint, along with 3D joint positions and velocities. Additionally, the corresponding 3D root translation vectors are concatenated into the pose representations to involve the trajectory of motion.
Previous group dance methods~\cite{le2023music}, which generate the whole group at once, cannot deal with the increasing number of dancers and can only create group sequences up to a pre-defined number of dancers, due to the vast complexity of the architecture. In contrast, we aim to generate group dance with an unlimited number of dancers.

\begin{figure*}[ht]
    \centering
    \includegraphics[width=1\textwidth, keepaspectratio=true]{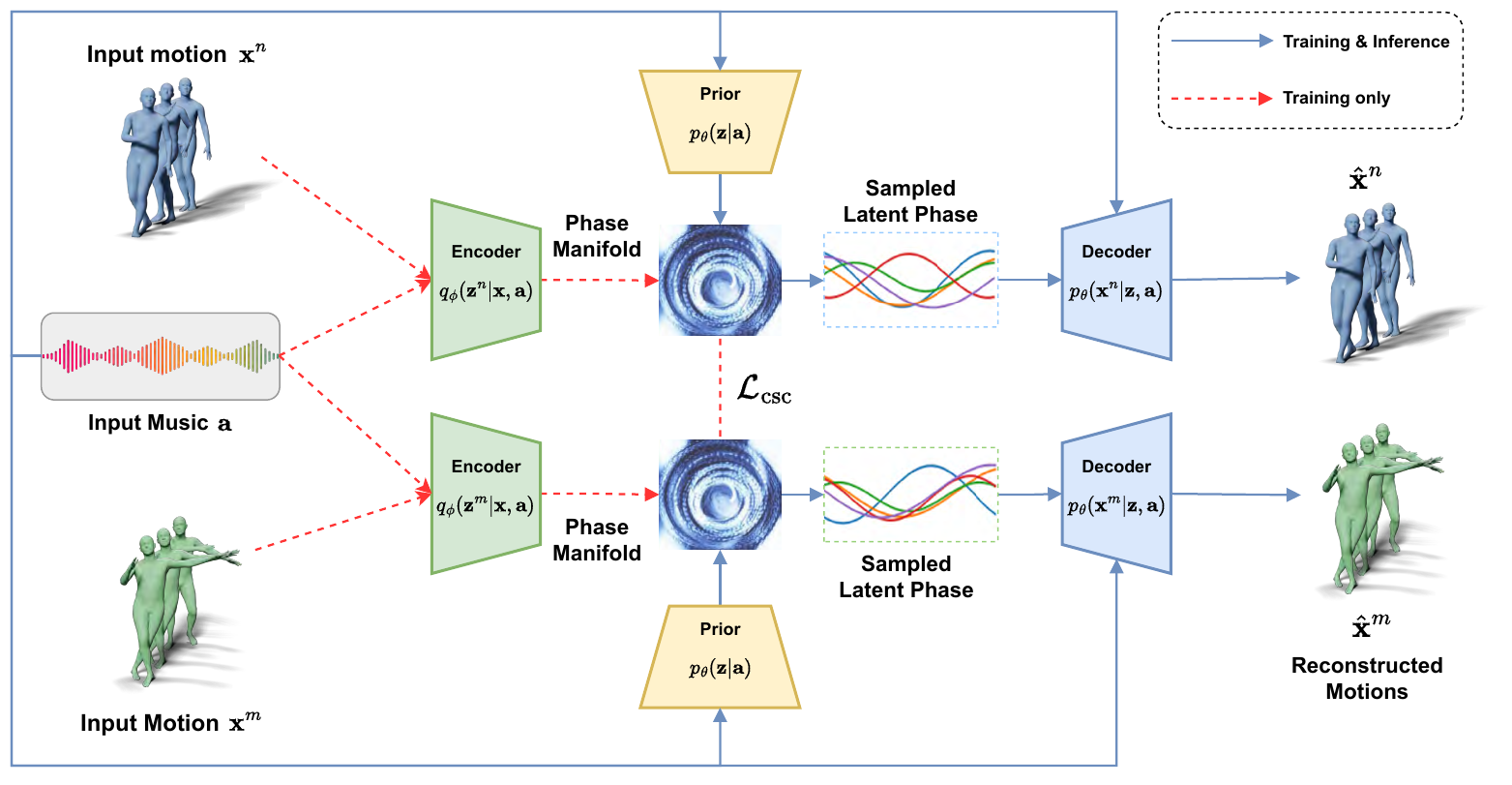}
    \vspace{-3ex}
    \caption{Overview of our Phase-conditioned Dance VAE (PDVAE) for scalable group dance generation. It consists of an Encoder, a Prior, and a Decoder network.  During training, we encode the corresponding motion and music inputs into a latent phase manifold, which is variational and parameterized by the frequency domain parameters of periodic functions. The latent phases can be sampled from the manifold and then decoded back to the original data space to obtain new motions.  The consistency loss $\mathcal{L}_{\text{csc}}$ is further imposed to constrain the manifold to be consistently unified for dancers that belong to the same group. At inference stage, only the Prior and the Decoder are used to synthesize group dances efficiently.  }
    \vspace{-1ex}
    \label{fig:CVAE}
\end{figure*}

\subsection{Phase-conditioned Dance VAE}
\label{sec:phase_cvae}
Our goal is to learn a continuous manifold such that the motion can be generated by sampling from this learned manifold. We assume that although different dancers within the same group may present visually distinctive movements, the properties of their motions, such as timing, periodicity, or temporal alignment are intrinsically similar. Drawing inspiration from~\cite{starke2022deepphase}, we aim to learn a generative phase representation for each group of dancer in order to synthesize their motion indefinitely. 
Our generative model is built upon the conditional Variational Autoencoder architecture~\cite{sohn2015cvae}, thanks to its diverse generation capability and fast sampling speed. However, instead of directly encoding the data into a Gaussian latent distribution as in common VAE approaches~\cite{petrovich2021action,petrovich2022temos_t2m_vae,lee2019_dancing2music,ghorbani2023speechgesture_vae, li2021audio2gestures}, we model the latent variational distribution by the phase parameters extracted from the latent motion curve, which we call variational phase manifold. The latent phase manifold is well-structured and can well describe key characteristics of motion (such as its timing, local periodicity, and transition), which benefits learning motion features~\cite{starke2022deepphase}. 

The overview of our Phase-conditioned Dance VAE is illustrated in Figure~\ref{fig:CVAE}. Specifically, the model contains three main networks: an encoder $\mathcal{E}$ to capture the approximate posterior distribution conditioned on both motion and music $q_\phi(\mathbf{z}|\mathbf{x},\mathbf{a})$, a prior network $\mathcal{P}$ to learn the conditional prior given only the music $p_\theta(\mathbf{z}|\mathbf{a})$ , and a decoder $\mathcal{D}$ to learn to reconstruct the data from the latent distribution $p_\theta(\mathbf{x}|\mathbf{z},\mathbf{a})$. The new motion is generated by sampling the frequency-domain parameters predicted by the prior network, which is then passed through the decoder network to reconstruct the motion in the original data space. Furthermore, we adopt Transformer-based architecture~\cite{vaswani2017attention} in each network to effectively capture long-range dependencies and holistic context of the dance sequence. 

\subsubsection{Encoder.}
The encoder $\mathcal{E}$ is expected to take both the motion and music feature sequence as input, and produce a distribution over possible latent variables capturing the cross-modal relationship between them. 
To transform the joint input space into a learned phase manifold, we adopt the Transformer decoder architecture where the Cross-Attention mechanism~\cite{vaswani2017attention} is utilized to learn the relationship between the motion and the music. 
Accordingly, the output of the encoder is a batch of latent curves (i.e., the activation sequences per channel) that can particularly capture different spatial and temporal aspects of the motion sequence. However, instead of training the model to directly reconstruct the input motion from the extracted latent curves, we further enforce each channel of the latent space to have a periodic functional form (i.e., sinusoidal). This enables us to effectively learn a compact parameterization for each latent channel from a small set of parameters in the frequency domain.

\subsubsection{Generative Variational Phase Manifold.}
\label{sec:phase_manifold}
Here we focus on learning the periodicity and non-linear temporal alignment of the motion in the latent space. In particular, given the output latent curves from the encoder $\mathbf{L} = \mathcal{E}(x,a) \in \mathbb{R}^{D \times T}$ with $D$ is the number of desired phase channels to be extracted from the motion, we parameterize each latent curve in $\mathbf{L}$ using a sinusoidal function with amplitude ($\mathbf{A}$), frequency ($\mathbf{F}$), offset ($\mathbf{B}$) and phase shift ($\mathbf{S}$) parameters~\cite{starke2022deepphase}. To allow for variational phase manifold learning, we opt to predict two sets of parameters $\mathbf{\mu}_{\mathcal{E}} =\{\mathbf{\mu}^A; \mathbf{\mu}^F; \mathbf{\mu}^B; \mathbf{\mu}^S \}$ and $\mathbf{\sigma}_{\mathcal{E}} =\{\mathbf{\sigma}^A; \mathbf{\sigma}^F; \mathbf{\sigma}^B; \mathbf{\sigma}^S \}$, which corresponds to the mean and variance of $\mathbb{R}^{4D}$ dimensional Gaussian distribution:
\begin{equation}
\label{eq:posterior}
    q_\phi(\mathbf{z}|\mathbf{x},\mathbf{a}) = \mathcal{N}(\mathbf{z};\mathbf{\mu}_{\mathcal{E}}, \mathbf{\sigma}_{\mathcal{E}})
\end{equation}
To do so, we first apply differentiable Fast Fourier Transform (FFT) to each channel of the latent curve $\mathbf{L}$ and create the zero-indexed matrix of Fourier coefficients as $\mathbf{c}=FFT(\mathbf{L})$ with $\mathbf{c} \in \mathbb{C}^{D \times K+1}$, $K =\lfloor \frac{T}{2}\rfloor$. Correspondingly, we compute the per channel power spectrum $\mathbf{p} \in \mathbb{R}^{D \times K+1}$ as $\mathbf{p}_{i,j} = \frac{2}{N}|\mathbf{c}_{i,j}|^2$, where $i$ is the channel index and $j$ is the index for the frequency bands. Correspondingly, the distributional mean parameters of the periodic sinusoidal function are then calculated as follows:
\begin{equation}
    \mathbf{\mu}^A_i = \sqrt{\frac{2}{T}\sum_{j=1}^K \mathbf{p}_{i,j}}, \quad
    \mathbf{\mu}^F_i = \frac{\sum_{j=1}^K \mathbf{f}_j \cdot \mathbf{p}_{i,j}}{ \sum_{j=1}^K \mathbf{p}_{i,j}}, \quad 
    \mathbf{\mu}^B_i = \frac{\mathbf{c}_{i,0}}{T},
\end{equation}
where $\mathbf{f} = (0, \frac{1}{T},\dots,\frac{K}{T})$ is the frequencies vector. At the same time, the phase shift $\mathbf{S}$ is predicted using a fully-connected (FC) layer with two $\arctan$ activation as:
\begin{equation}
    \label{eq:fc_phase_shift}
    (s_y, s_x)  = \text{FC}(\mathbf{L}_i), \quad \mathbf{\mu}^S_i = \arctan(s_y,s_x),
\end{equation}

To predict the distributional variance of the phase amplitude and phase shift parameters $\{\mathbf{\sigma}^A, \mathbf{\sigma}^S\}$, We additionally apply a separate two-layer MLP network over each channel of the latent curves, similar to Equation~\ref{eq:fc_phase_shift}. The variational latent phase parameters are sampled by utilizing parameterization trick~\cite{kingma2013vae}, i.e., $\mathbf{A}\sim\mathcal{N}(\mathbf{\mu}^A,\mathbf{\sigma}^A)$ and $\mathbf{S}\sim\mathcal{N}(\mathbf{\mu}^S,\mathbf{\sigma}^S)$. In our experiments, we find that sampling the phase frequency $\mathbf{F}$ and offset $\mathbf{B}$ often produce unstable and non-coherent group movements. This might be because the frequency amplitudes of the dancers within the same group are likely to associate with the rhythmic pattern of the musical beats while the offsets capture their alignment, thereby should be consistent with each other. Therefore, we treat those parameters as deterministic by constraining their variance to zero.

Finally, the sampled set of phase parameters $\mathbf{z} = \{\mathbf{A};\mathbf{F};\mathbf{B};\mathbf{S}\}$ are used to reconstruct a parametric latent space consisting of multiple periodic curves to represent each intrinsic property of the motion by:
\begin{equation}
\label{eq:curve_construct}
    \hat{\mathbf{L}} = \mathbf{A} \cdot \sin (2\pi \cdot (\mathbf{F}\cdot\mathcal{T} - \mathbf{S})) + \mathbf{B}
\end{equation}
where $\mathcal{T}$ is a known time window series obtained by evenly spacing the timesteps from $0$ to $T$. Intuitively this curve construction procedure can be viewed as a "quantization" layer to enforce the network to learn to represent the motion features in the frequency domain, which is useful in representing different aspects of human motion such as their timing and periodicity. In the last step, a decoder is utilized to reconstruct the original motion signals from the set of parametric latent curves.


\textbf{Decoder.}  To decode the latent space into the original motion space, previous works~\cite{petrovich2022temos_t2m_vae, petrovich2021action} have to use a sinusoidal positional encoding sequence with duration $T$ as the proxy input to the sequence decoder. This is because their latent space is only formed by single latent vectors following a Gaussian distribution, which cannot span the time dimension. However, we observe that it usually results in unstable and inconsistent movements, as the proxy sequence is generic and usually contains less meaningful information for the decoder. Meanwhile, our method does not suffer from this problem as our latent space is built on multiple curves that can represent the motion information through time, thanks to the phase parameters. 
Subsequently, our decoder $\mathcal{D}$ is based on Transformer decoder architecture that takes the constructed parametric latent curve, as well as the music features as inputs, to reconstruct the corresponding dance motions.
Here, we also utilize the cross-attention model~\cite{vaswani2017attention} where we consider the sequence of and music features as key and value along with the sampled latent curves as the query. The output of the decoder is a sequence of $T$ vectors in $\mathbb{R}^D$, which is then projected back to the original motion dimensions through a linear layer, to obtain the reconstructed outputs $\hat{\mathbf{x}}=p_\theta(\mathbf{x}|\mathbf{z},\mathbf{a})$. We additionally employ a global trajectory predictor to predict the global translation of the root joint based on the generated local motions~\cite{he2022nemf,li2021hm_vae}, in order to avoid intersection problems between dancers. We provide details of our global predictor in the supplemental.

\subsubsection{Prior Network.}
Since the ground-truth motion is generally inaccessible at test time (i.e., we only have access to the music), we also need to learn a prior $\mathcal{P}$ to match the posterior distribution of motion from which the latent phase can be sampled. Specifically, We follow the procedure similar to Section \ref{sec:phase_manifold} (Equation~\ref{eq:posterior}-\ref{eq:curve_construct}) to predict the Gaussian distribution conditioned on the music sequence $\mathbf{a}$, which is then used for sampling the latent phases:
\begin{equation}
\label{eq:prior}
    p_\theta(\mathbf{z}|\mathbf{a}) = \mathcal{N}(\mathbf{z};\mathbf{\mu}_{\mathcal{P}}, \mathbf{\sigma}_{\mathcal{P}})
\end{equation}
where a Transformer encoder is used to encode the input conditioning music sequence and predict the corresponding $\mathbf{\mu}_{\mathcal{P}}$ and $\mathbf{\sigma}_{\mathcal{P}}$.
We implement the prior network similarly to the encoder network, however, we use self-attention mechanism~\cite{vaswani2017attention} to capture the global music context. Learning the conditional prior is crucial for the conditional VAE to generalize to diverse types of music and motion. Intuitively speaking, each latent variable $\mathbf{z}$ is expected to represent possible dance motions $\mathbf{x}$ conforming to the music context $\mathbf{a}$. Therefore, the prior should be able to encode different latent distributions given different musics.
 

\subsection{Training} 
During training, we consider the following variational lower bound~\cite{sohn2015cvae} to mainly train our dance generation VAE model:
\begin{equation}
\log p_\theta(\mathbf{x}|\mathbf{a}) \geq \mathbb{E}_{q_\phi} \left[ \log p_\theta(\mathbf{x}|\mathbf{z},\mathbf{a}) \right] - D_{\text{KL}}(q_\phi(\mathbf{z}|\mathbf{x},\mathbf{a}) \Vert p_\theta(\mathbf{z}|\mathbf{a}))
\end{equation}
In practice, we apply the conditional VAE loss as similar to~\cite{petrovich2022temos_t2m_vae}, which is defined as the weighted sum $\mathcal{L}_{\text{cvae}} = \mathcal{L}_{\text{rec}} + \lambda_{\text{KL}}\mathcal{L}_{\text{KL}}$. In particular, the reconstruction term $\mathcal{L}_{\text{rec}}$ measures the motion reconstruction error given the decoder output (via a smooth-L1 loss). The KL divergence term $\mathcal{L}_{\text{KL}}$ compares the divergence $D_{\text{KL}}$ between the posterior and the prior distribution to enforce them to be close to each other.

The conditional VAE objective above is calculated for each dancer separately and cannot capture the correlation between dancers within a group. Therefore, it is essential to impose consistency among dancers and avoid strange effects such as unsynchronized dance. To this end, we propose a group consistency loss by 
enforcing the latent phase manifold to be similar for the same group, given the input music. Specifically, we first calculate the phase manifold features based on the frequency domain parameters as follows:
\begin{equation}
    \mathbf{P}_{2i-1} = \mathbf{A}_i\sin(2\pi \cdot \mathbf{S}_i), \qquad \mathbf{P}_{2i} = \mathbf{A}_i\cos(2\pi\cdot \mathbf{S}_i)
\end{equation}
where $\mathbf{P}\in\mathbb{R}^{2D}$ is the phase manifold vector that encodes the spatial-temporal information of the motion state. The phase feature may look similar to the positional encodings of transformers~\cite{vaswani2017attention} in the sense that both use multi-resolution sinusoidal functions. However, the phase feature is much richer in terms of representation capacity since it learns to embed the spatial (body joints) and temporal (positions in time) information of the motion curves, whereas the positional encodings only encode the position of words. Finally, our consistency objective is to constrain phase manifold between dancers within a group to be as close as possible to each other, which is formulated as:
\begin{equation}
    \mathbf{\mathcal{L}}_{\text{csc}} =  D_{\text{KL}}(q_\phi(\mathbf{z}|\mathbf{x}^m,\mathbf{a}) \Vert (q_\phi(\mathbf{z}|\mathbf{x}^n,\mathbf{a}) ) + \Vert \mathbf{P}^m - \mathbf{P}^n\Vert^2_2
\end{equation}
where the first term encourages the network to map different dancers belonging to the same group ($\mathbf{x}^m$ and $\mathbf{x}^n$) into the same set of distributional phase parameters while the second term penalizes the discrepancy in their corresponding phase manifolds. In general, this loss is applied to ensure every dancer is embedded into a single unified manifold that can effectively represent their corresponding group. To summarize, our total training loss is defined as the combination of the VAE loss and the consistency loss $\mathcal{L} = \mathbf{\mathcal{L}}_{\text{cvae}} + \lambda_{\text{csc}}\mathbf{\mathcal{L}}_{\text{csc}}$.

For testing, we can efficiently generate motions for an unlimited number of dancers by indefinitely drawing samples from the learned continuous group-consistent phase manifold. It is noteworthy that for inference, we only need to process the prior network once to obtain the latent distribution. To generate a new motion, we can sample from this latent (Gaussian) distribution and use the decoder to decode it back to the motion space. This approach is much more efficient and has significantly higher scalability than previous approaches~\cite{le2023controllableGCD,le2023music} that is limited by the number of dancers processed simultaneously by the entire network.

\section{Experiments}
\subsection{Implementation Details}
Our model was trained on 4 NVIDIA V100 GPUs using Adam optimizer with a fixed learning rate of $10^{-4}$ and a mini-batch size of 32 per GPU. For training losses, the weights are empirically set to $\lambda_{\text{KL}} = 5\times 10^{-4}$ and $\lambda_{\text{csc}} = 10^{-4}$, respectively. The Transformer encoders and decoders consist of 5 layers for both encoder, decoder, and prior Network with 512-dimensional hidden units. Meanwhile, the number of latent phase channels is set to 256. To further capture the periodic nature of the phase feature, we also use Siren activation following the initialization scheme as in~\cite{sitzmann2020siren}. This can effectively model the periodicity inherent in the motion data, and thus can benefit motion synthesis~\cite{shi2023phasemp}. 

\subsection{Experimental Settings}
\label{sec:metric}

\subsubsection{Dataset}

In our experiments, we utilize the AIOZ-GDance~\cite{le2023music} and AIST-M~\cite{yao2023dance} datasets. AIOZ-GDance is the largest music-driven dataset focusing on group dance, encompassing paired music and 3D group motions extracted from in-the-wild videos through a semi-automatic process. This dataset spans 7 dance styles and 16 music genres. For consistency, we adhere to the training and testing split outlined in~\cite{le2023music} during our experiments.

\subsubsection{Evaluation Protocol}
We employ several metrics to assess the quality of individual dance motions, including Frechet Inception Distance (FID)~\cite{heusel2017ganfid,li2021AIST++}, Motion-Music Consistency (MMC)~\cite{li2021AIST++}, and Generation Diversity (GenDiv)~\cite{Dance_Revolution, lee2019_dancing2music,li2021AIST++}, along with the Physical Foot Contact score (PFC)~\cite{tseng2022edge_diffusion}. Specifically, the FID score gauges the realism of individual dance movements concerning the ground-truth dance. MMC assesses the matching similarity between motion and music beats, reflecting how well-generated dances synchronize with the music's rhythm. GenDiv is computed as the average pairwise distance of kinetic features among motions~\cite{onuma2008fmdistance}. PFC evaluates the physical plausibility of foot movements by determining the agreement between the acceleration of the character's center of mass and the foot's velocity.

In assessing the quality of group dance, we adopt three metrics outlined in~\cite{le2023music}: Group Motion Realism (GMR), Group Motion Correlation (GMC), and Trajectory Intersection Frequency (TIF). Broadly, GMR gauges the realism of generated group motions in comparison to ground-truth data, employing Frechet Inception Distance on extracted group motion features. GMC evaluates the synchronization among dancers within the generated group by computing their cross-correlation. TIF quantifies the frequency of collisions among the generated dancers during their dance movements.


\subsubsection{Baselines}  
Our method is subjected to comparison with various recent techniques in music-driven dance generation, namely FACT~\cite{li2021AIST++}, Transflower~\cite{perez2021_transflower}, and EDGE~\cite{tseng2022edge_diffusion}. These approaches are adapted for benchmarking within the context of group dance generation, as outlined in~\cite{le2023music}, considering that their original designs were tailored for single-dance scenarios. Additionally, our evaluation includes a comparison with GDanceR~\cite{le2023music}, GCD~\cite{le2023controllableGCD}, and DanY~\cite{yao2023dance}. All of the mentioned works are specifically designed for the generation of group choreography.

\vspace{-3ex}
\begin{table}[h]
\centering
\caption{Performance comparison.}
\resizebox{\linewidth}{!}{
\setlength{\tabcolsep}{0.4 em} 
{\renewcommand{\arraystretch}{1.2}
\begin{tabular}{l|lc|cccc|ccc}
\hline
\multicolumn{1}{c|}{\textbf{Dataset}} &\multicolumn{2}{c|}{\textbf{Method}} & \multicolumn{1}{c|}{FID$\downarrow$} & \multicolumn{1}{c|}{MMC$\uparrow$} & \multicolumn{1}{c|}{GenDiv$\uparrow$} &   {PFC}$\downarrow$ & \multicolumn{1}{c|}{GMR$\downarrow$} & \multicolumn{1}{c|}{GMC$\uparrow$} & TIF$\downarrow$ \\   \hline
\multirow{7}{*}{\textbf{AIOZ-GDANCE}~\cite{le2023music}} &\multicolumn{2}{l|}{FACT~\cite{li2021AIST++}} & \multicolumn{1}{c|}{56.20} & \multicolumn{1}{c|}{0.222} & \multicolumn{1}{c|}{8.64} & 3.52 & \multicolumn{1}{c|}{101.52} & \multicolumn{1}{c|}{62.68} & 0.321 \\ 
&\multicolumn{2}{l|}{  {Transflower {\cite{perez2021_transflower}}}} & \multicolumn{1}{c|}{37.73} & \multicolumn{1}{c|}{0.217} & \multicolumn{1}{c|}{8.74} & 3.07 & \multicolumn{1}{c|}{81.17} & \multicolumn{1}{c|}{60.78} & 0.332 \\ 
&\multicolumn{2}{l|}{  {EDGE {\cite{tseng2022edge_diffusion}}}} & \multicolumn{1}{c|}{31.40} & \multicolumn{1}{c|}{0.264} & \multicolumn{1}{c|}{9.57} & 2.63 & \multicolumn{1}{c|}{63.35} & \multicolumn{1}{c|}{61.72} & 0.356 \\ 
\cline{2-10}
&\multicolumn{2}{l|}{GDANCER~\cite{le2023music}} & \multicolumn{1}{c|}{43.90} & \multicolumn{1}{c|}{0.250} & \multicolumn{1}{c|}{9.23} & 3.05 & \multicolumn{1}{c|}{51.27} & \multicolumn{1}{c|}{79.01} & 0.217  \\

&\multicolumn{2}{l|}{GCD~\cite{le2023controllableGCD}} & \multicolumn{1}{c|}{{31.16}} & \multicolumn{1}{c|}{0.261} & \multicolumn{1}{c|}{10.87} & 2.53 &\multicolumn{1}{c|}{31.47} & \multicolumn{1}{c|}{80.97} & {0.167}  \\

\cline{2-10}
&\multicolumn{2}{l|}{\textbf{PDVAE (ours)}} & \multicolumn{1}{c|}{\textbf{31.01}} & \multicolumn{1}{c|}{\textbf{0.271}} & \multicolumn{1}{c|}{\textbf{10.98}} & \textbf{2.33} & \multicolumn{1}{c|}{\textbf{30.08}} & \multicolumn{1}{c|}{\textbf{84.52}} & \textbf{0.163}  \\
\hline \hline
\multirow{7}{*}{\textbf{AIST-M}~\cite{yao2023dance}} 
&\multicolumn{2}{l|}{GDANCER~\cite{le2023music}} & \multicolumn{1}{c|}{52.90} & \multicolumn{1}{c|}{0.222} & \multicolumn{1}{c|}{6.52} & 1.93 & \multicolumn{1}{c|}{65.13} & \multicolumn{1}{c|}{60.56} & 0.121  \\

&\multicolumn{2}{l|}{GCD~\cite{le2023controllableGCD}} & \multicolumn{1}{c|}{{35.36}} & \multicolumn{1}{c|}{0.245} & \multicolumn{1}{c|}{10.97} & 1.52 &\multicolumn{1}{c|}{42.52} & \multicolumn{1}{c|}{72.15} & 0.083  \\

&\multicolumn{2}{l|}{DanY~\cite{yao2023dance}} & \multicolumn{1}{c|}{40.25} & \multicolumn{1}{c|}{0.240} & \multicolumn{1}{c|}{11.40} & 1.65 & \multicolumn{1}{c|}{50.29} & \multicolumn{1}{c|}{63.53} & 0.137 \\
\cline{2-10}
&\multicolumn{2}{l|}{\textbf{PDVAE (ours)}} & \multicolumn{1}{c|}{\textbf{31.49}} & \multicolumn{1}{c|}{\textbf{0.257}} & \multicolumn{1}{c|}{\textbf{11.81}} & \textbf{1.42} & \multicolumn{1}{c|}{\textbf{41.24}} & \multicolumn{1}{c|}{\textbf{78.64}} & \textbf{0.076}  \\
\hline
\end{tabular}
}}
    \label{tab:main}
\end{table}
\vspace{-6ex}

\subsection{Experimental Results}
\label{sec:quantitative}
\subsubsection{Quality Comparison}
\begin{figure*}[h] 
   \centering
\resizebox{\linewidth}{!}{
\setlength{\tabcolsep}{2pt}
\begin{tabular}{cccccc}
\rotatebox[origin=l]{90}{\hspace{0.1cm} \textbf{GDanceR}} &
\shortstack{\includegraphics[width=0.33\linewidth]{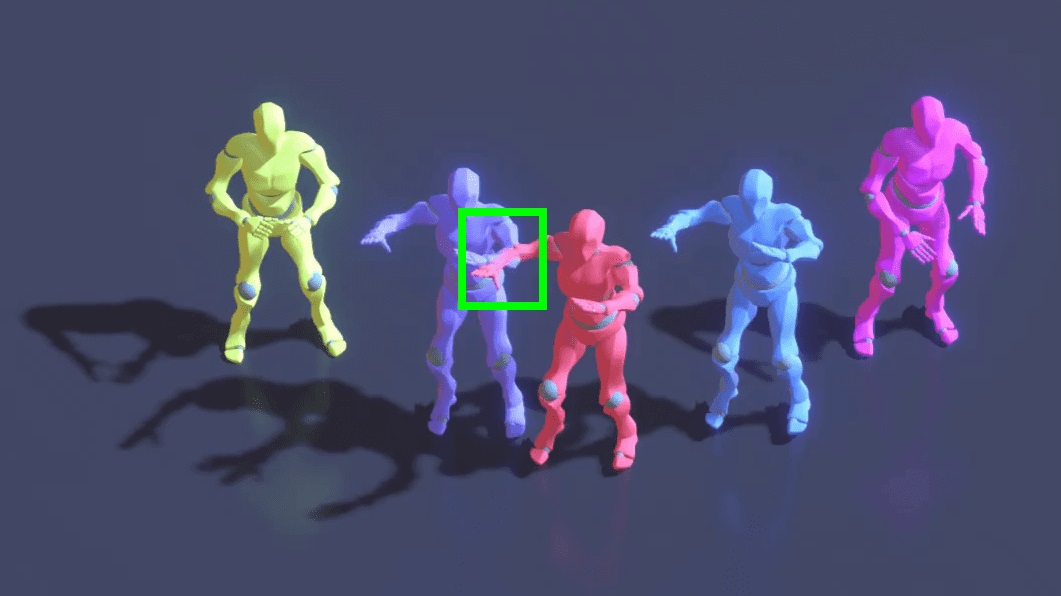}}&
\shortstack{\includegraphics[width=0.33\linewidth]{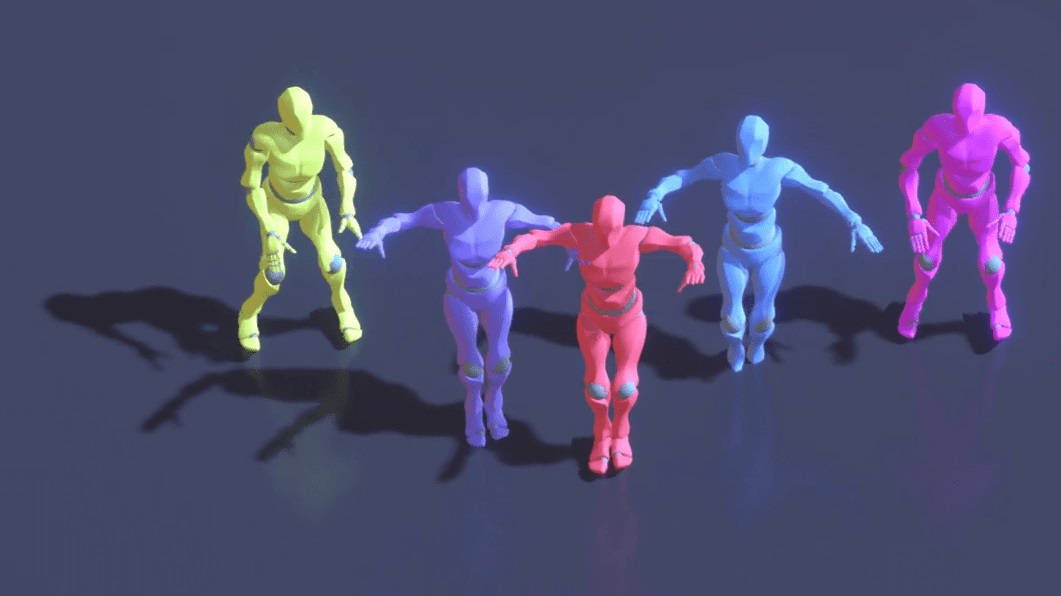}}&
\shortstack{\includegraphics[width=0.33\linewidth]{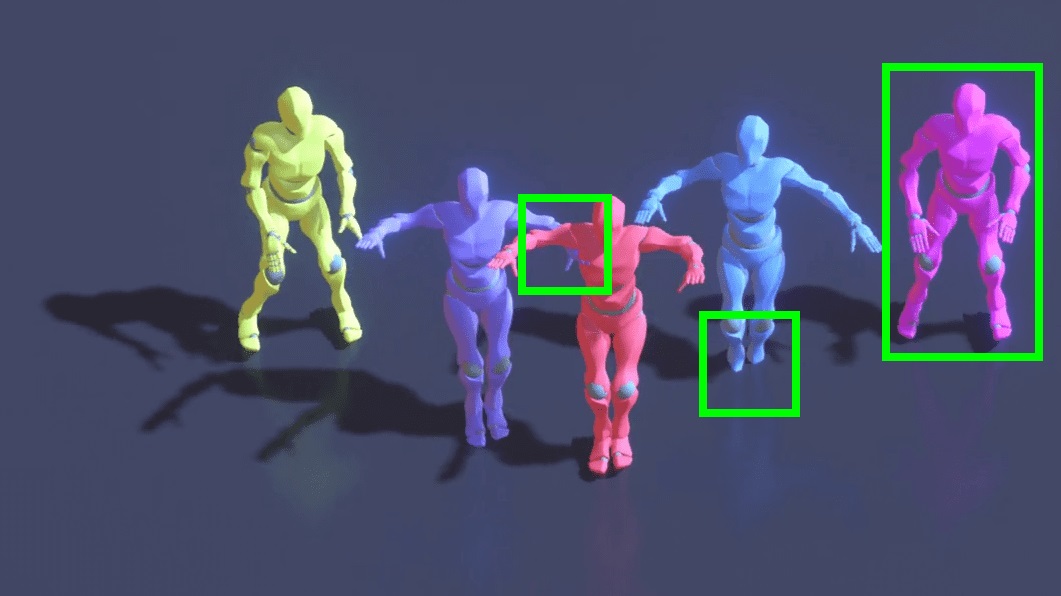}}&
\shortstack{\includegraphics[width=0.33\linewidth]{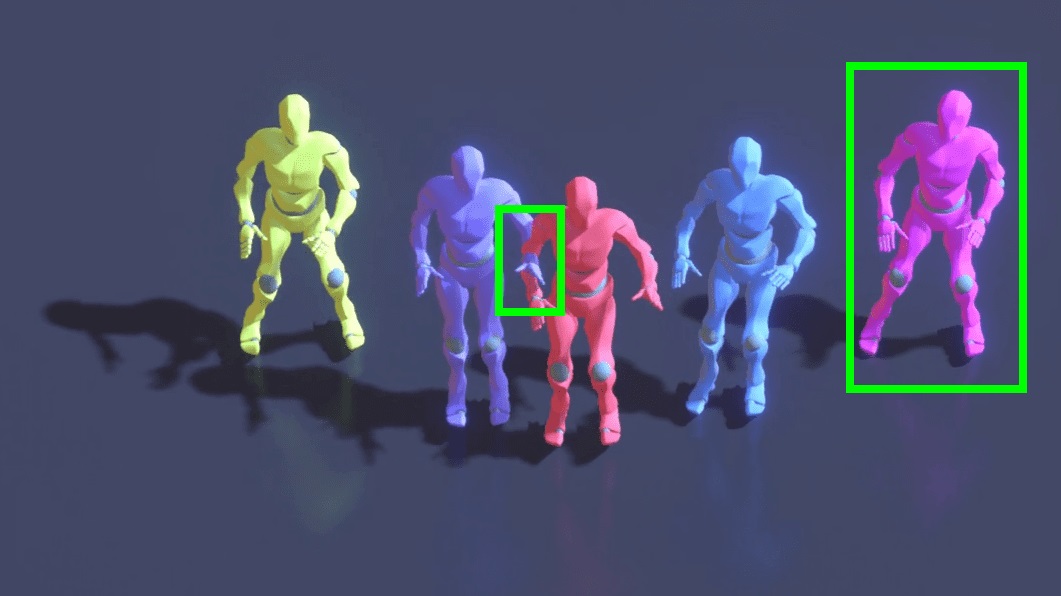}}\\[1pt]
\rotatebox[origin=l]{90}{\hspace{0.35cm} \textbf{GCD}} &
\shortstack{\includegraphics[width=0.33\linewidth]{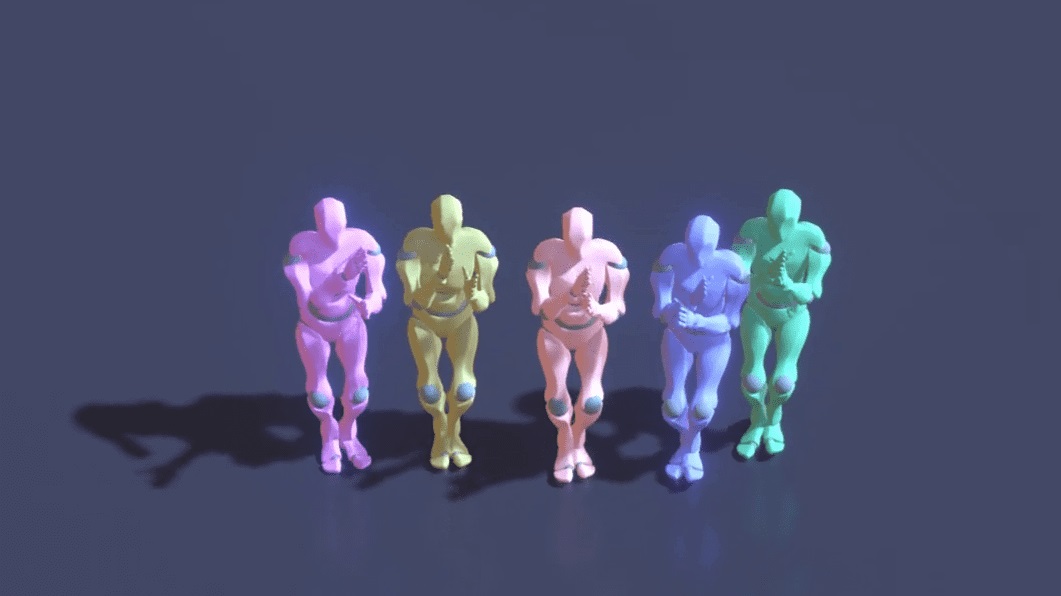}}&
\shortstack{\includegraphics[width=0.33\linewidth]{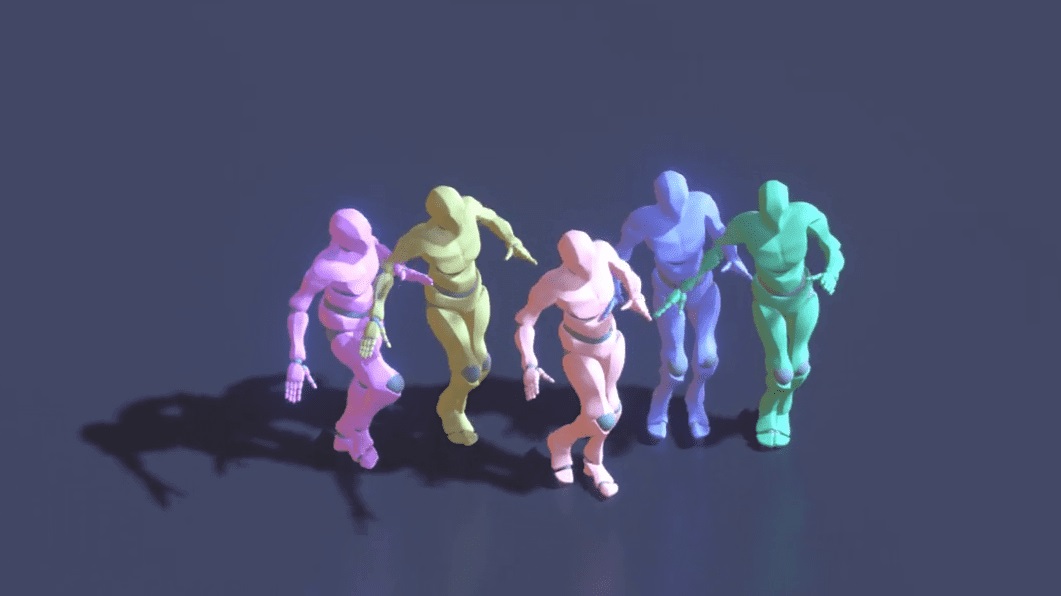}}&
\shortstack{\includegraphics[width=0.33\linewidth]{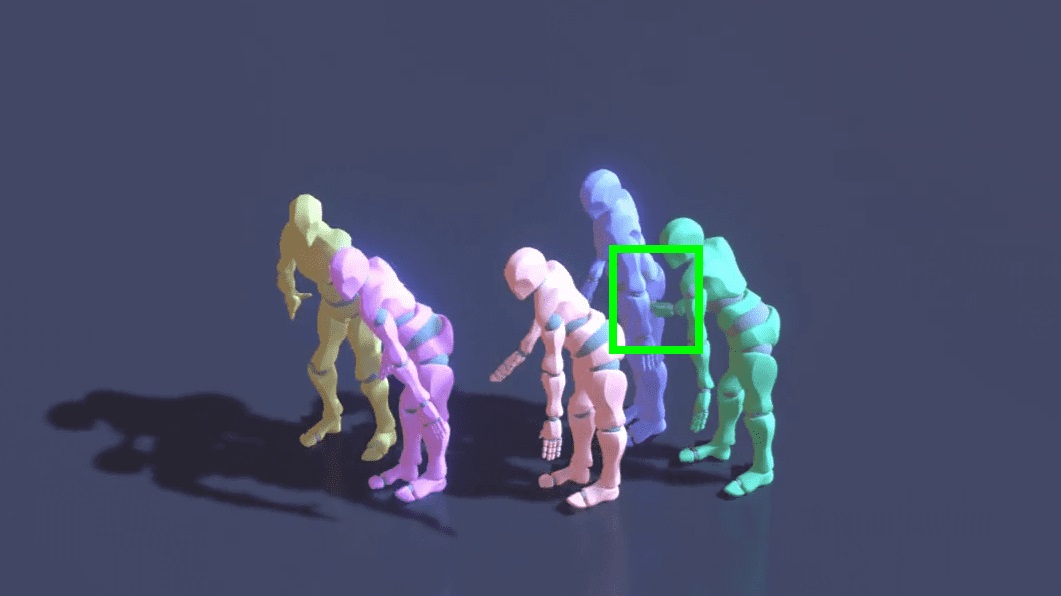}}&
\shortstack{\includegraphics[width=0.33\linewidth]{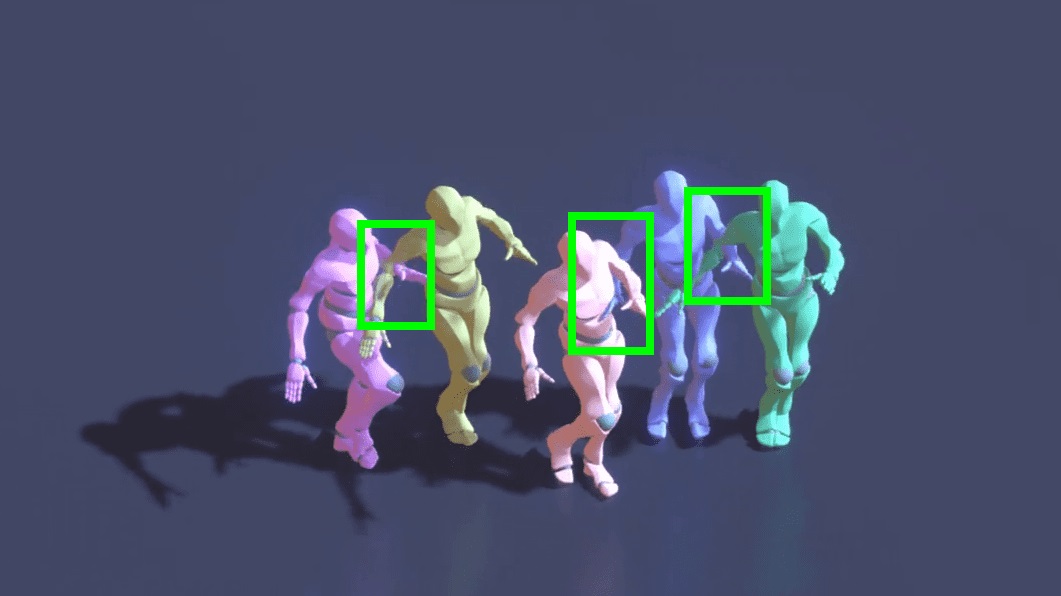}}\\[1pt]
\rotatebox[origin=l]{90}{\hspace{0.5cm} \textbf{Ours}} &
\shortstack{\includegraphics[width=0.33\linewidth]{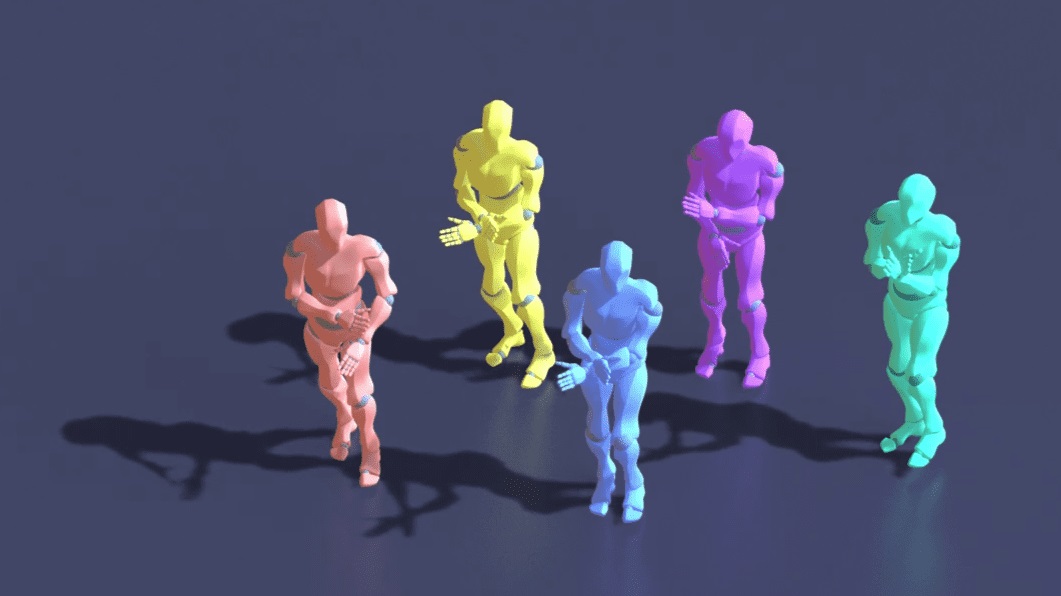}}&
\shortstack{\includegraphics[width=0.33\linewidth]{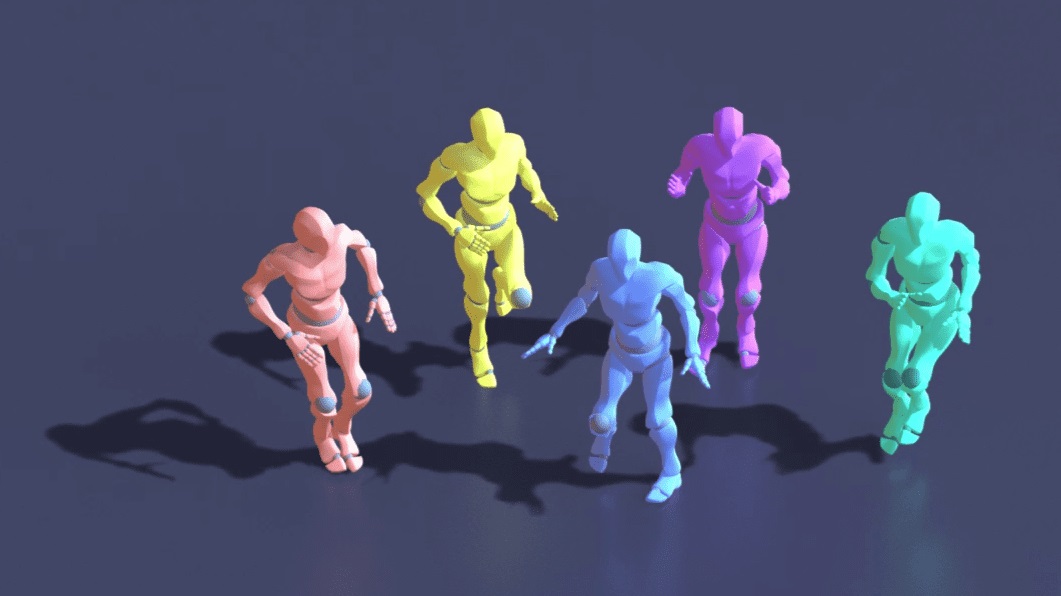}}&
\shortstack{\includegraphics[width=0.33\linewidth]{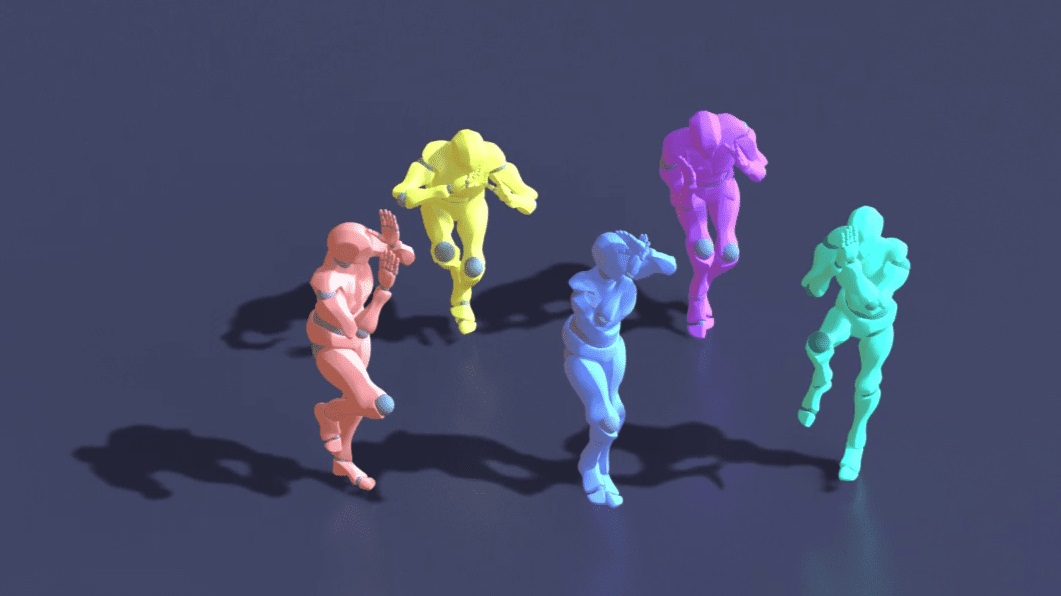}}&
\shortstack{\includegraphics[width=0.33\linewidth]{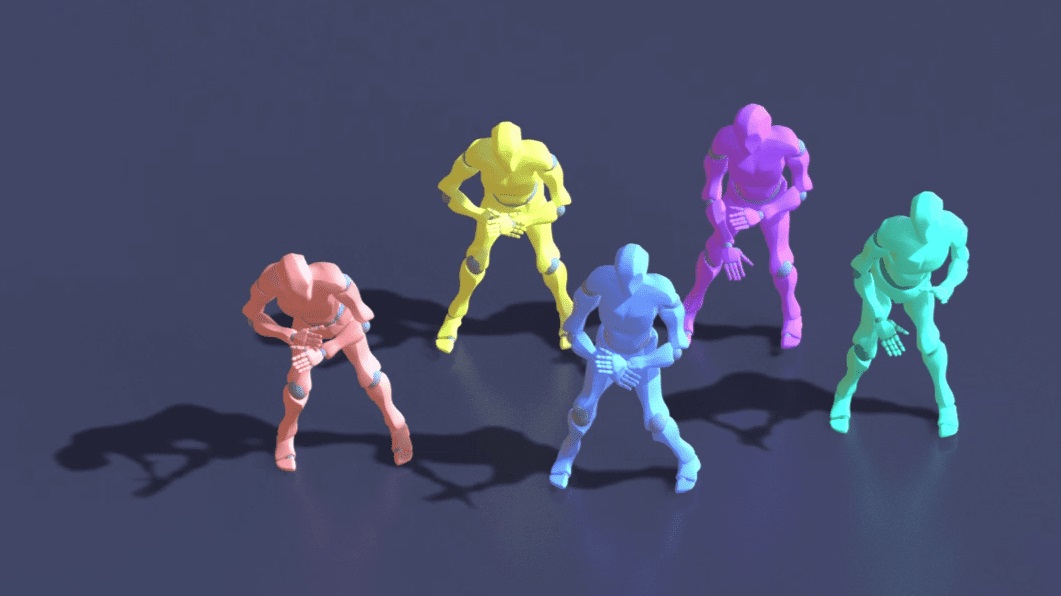}}\\[1pt]

\end{tabular}
}
    \caption{Visualization of a dancing sample between different methods. GDanceR displays monotonous, repetitive, or sinking dance motions. GCD exhibits more divergence in dance motions, yet dancers may intersect since their optimization does not address this issue explicitly. Blue boxes mark these issues. In contrast, our manifold-based solution ensures the divergence of dancing motions, while the phase motion path demonstrates its effectiveness in addressing floating and crossing issues in group dances. 
    \label{fig:Comapre}
    } 
\end{figure*}

Table~\ref{tab:main} presents a comparison among the baselines FACT~\cite{li2021AIST++}, Transflower~\cite{perez2021_transflower}, EDGE~\cite{tseng2022edge_diffusion}, GDanceR~\cite{le2023music}, GCD~\cite{le2023controllableGCD}, and our proposed manifold-based method. The results clearly demonstrate that our model almost outperforms all baselines across all evaluations on two datasets AIOZ-GDANCE~\cite{le2023music} and AIST-M~\cite{yao2023dance}. We observe that recent diffusion-based dance generation models such as EDGE or GCD yield competitive performance on both single-dance metrics (FID, MMC, GenDiv, and PFC) and group dance metrics (GMR, GMC, and TIF). However, due to limitations in their training procedures, they still struggle when dealing with generating multiple dancing motions in the context of a high quantity of dancers, as shown by their low performance compared to our proposal. This result implies that our method successfully preserves the performance of dancing motions when the number of dancers is increased. Besides, Figure~\ref{fig:Comapre} also shows that our proposed methods outperform other state-of-the-art methods such as GDanceR or GCD in dealing with monotonous, repetitive, sinking, and overlapping dance motions.



\vspace{-2ex}
\vspace{-6ex}

\begin{table}[h]
\centering
\caption{Performance of group dance generation methods when we increase the number of generated dancers. The experiments are done with common consumer GPUs with 4GB memory. (N/A means models could not run due to inadequate memory footprint)}

\resizebox{\linewidth}{!}{
\setlength{\tabcolsep}{0.3 em} 
{\renewcommand{\arraystretch}{1.2}
\begin{tabular}{c|c|c|c|c|c|c|c}
  \hline
\textbf{\begin{tabular}[c]{@{}c@{}}\#Generated\\ Dancers\end{tabular}} & \textbf{Method}  & \textbf{FID}$\downarrow$ & \textbf{MMC}$\uparrow$ & \textbf{GenDiv}$\uparrow$ & \textbf{GMR}$\downarrow$ & \textbf{GMC}$\uparrow$ & \textbf{TIF}$\downarrow$\\   \hline
\multirow{3}{*}{\textbf{5}} 
 & GDanceR~\cite{le2023music} & 44.19 & 0.249 & 8.99 & 55.05  & 78.72 & 0.218 \\
 & GCD~\cite{le2023controllableGCD} & 35.08 & 0.264 & 9.92 & 38.43 & 81.44 & 0.168 \\ 
 & PDVAE (ours) & \textbf{31.35}  & \textbf{0.268}  & \textbf{10.05}  & \textbf{32.58}  & \textbf{84.56} & \textbf{0.161} \\   \hline\hline
\multirow{3}{*}{\textbf{10}} 
 & GDanceR~\cite{le2023music} & 57.93  & 0.225  & 7.63  & 73.32  & 72.79  & 0.386 \\
 & GCD~\cite{le2023controllableGCD} & N/A (Memory) & N/A (Memory) & N/A (Memory)  & N/A (Memory) & N/A (Memory) & N/A (Memory)\\ 
 & PDVAE (ours) & \textbf{32.19}  & \textbf{0.269}  & \textbf{8.99}  & \textbf{34.32}  & \textbf{86.96} & \textbf{0.193} \\   \hline\hline
 \multirow{3}{*}{\textbf{100}} 
 & GDanceR~\cite{le2023music} & N/A (Memory)  & N/A (Memory)  & N/A (Memory)  & N/A (Memory)  & N/A (Memory)  & N/A (Memory) \\
 & GCD~\cite{le2023controllableGCD} & N/A (Memory)  & N/A (Memory)  & N/A (Memory)  & N/A (Memory)  & N/A (Memory) & N/A (Memory)\\ 
 & PDVAE (ours) & \textbf{30.97}  & \textbf{0.268}  & \textbf{8.76} & \textbf{38.13}  & \textbf{85.73} & \textbf{0.222} \\   \hline\hline
 
\end{tabular}
}}
\vspace{2ex}
\label{tab:n_dancers}
\vspace{-1ex}
\end{table}

\subsubsection{Scalable Generation Analysis}
Table~\ref{tab:n_dancers} illustrates the performance of different group dance generation methods (GCD, GDancer, and Ours) when generating dance movements with an increasing number of dancers. When the number of dancers is increased to 10, GCD appears to run out of memory, which is also observed in GDanceR when the number of dancers increases to 100. Regardless of the number of dancers, our method consistently achieves stable and competitive results. This implies that our proposed method successfully addresses the scalability issue in group dance generation without compromising the overall performance of each individual dance motion.
Figure~\ref{fig:memoryUsage} illustrates the memory consumption to generate dance motions in groups for each method. Noticeably, our proposal still achieves the highest performance while consuming immensely fewer resources required for generating group dance motions (See Figure~\ref{fig:ScaleVis} for illustrations). This, again, indicates that our method successfully breaks the barrier of limited generated dancers by using the manifold.

\subsubsection{Ablation Analysis} Table~\ref{tab:ComponentAnalysis} presents the performance improvements achieved through the integration of consistency loss and phase manifold. Additionally, we showcase the effectiveness of our proposed approach across three different backbones: Transformer~\cite{vaswani2017attention,li2021AIST++}, LSTM~\cite{lstm}, and CNN~\cite{zhuang2020_music2dance}. Evaluation metrics including FID, GMR, and GMC are utilized. The results indicate that the absence of consistency loss leads to an increase in GMR and a decrease in GMC, suggesting a significant enhancement in the realism and correlation of group dance motions facilitated by the inclusion of the proposed objective. Meanwhile, with out the phase manifold, the model exhibits remarkably higher scores in both the FID and GMR metrics, suggesting that phase manifold can effectively improve the distinction in dance motions while maintaining the realism of group dances, even when the number of dancers in a group is large. In comparing three backbones—Transformer, LSTM, and CNN—we have observed that the chosen Transformer backbone achieved the best results compared to LSTM or CNN. 


\vspace{-2ex}
\begin{table}[h]
\begin{minipage}[b]{0.48\linewidth}
\centering
\resizebox{\linewidth}{!}{
\setlength{\tabcolsep}{0.2 em} 
{\renewcommand{\arraystretch}{1.2}
\begin{tabular}{lc|c|cc}
\hline
\multicolumn{2}{c|}{\textbf{Method}} & \multicolumn{1}{c|}{FID$\downarrow$}  & \multicolumn{1}{c|}{GMR$\downarrow$} & \multicolumn{1}{c}{GMC$\uparrow$} \\  \hline
\multicolumn{2}{l|}{\textbf{Ours}} & \multicolumn{1}{c|}{31.01} & \multicolumn{1}{c|}{30.08} & \multicolumn{1}{c}{84.52} \\ 
\multicolumn{2}{l|}{Ours w/o Consistency Loss} & \multicolumn{1}{c|}{35.35} & \multicolumn{1}{c|}{57.63} & \multicolumn{1}{c}{66.72}\\ 
\multicolumn{2}{l|}{Ours w/o Phase Manifold} & \multicolumn{1}{c|}{41.78} & \multicolumn{1}{c|}{45.32} & \multicolumn{1}{c}{77.93}\\ 
\hline
\multicolumn{2}{l|}{Ours w LSTM}  & \multicolumn{1}{c|}{41.29} & \multicolumn{1}{c|}{47.47} & \multicolumn{1}{c}{71.82} \\ 
\multicolumn{2}{l|}{Ours w CNN} & \multicolumn{1}{c|}{36.99} & \multicolumn{1}{c|}{44.94} & \multicolumn{1}{c}{75.77} \\ 
 \hline
\end{tabular}
}}
\vspace{2ex}
    \caption{Module contribution and loss analysis.}
    \label{tab:ComponentAnalysis}
\end{minipage}
\hfill
\begin{minipage}[b]{0.48\linewidth}
\centering
\includegraphics[width=\textwidth, keepaspectratio=true]{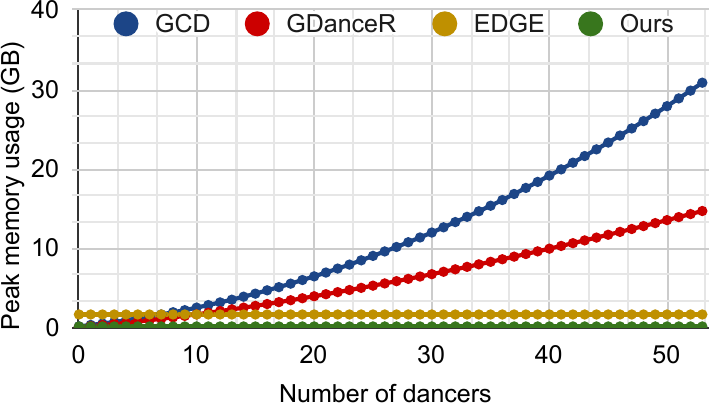}
\vspace{-3ex}
\captionof{figure}{Memory usage vs. number of dancers in different dance generators.
}
\label{fig:memoryUsage}

\end{minipage}
\end{table}
\vspace{-8ex}

\subsection{User Study}

User studies are vital for evaluating generative models, as user perception is pivotal for downstream applications; thus, we conducted two studies with around 70 participants each, diverse in background, with experience in music and dance, aged between 20 to 40, consisting of approximately 47\% females and 53\% males, to assess the effectiveness of our approach in group choreography generation.




In the user study, we aim to assess the realism of sample outputs with more and more dancers. Specifically, participants assign scores ranging from 0 to 10 to evaluate the realism of each dance clip with 2 to 10 dancers. Figure~\ref{fig:UserStudy2} shows that, across all methods, the more the number of dancers is increased, the lower the realism is found. However, the drop in realism of our proposed method is the least compared to GCD and GDanceR. The results indicate our method's good performance compared to other baselines when the number of dancers increases. 

\vspace{-2ex}
\begin{table}[h]
\begin{minipage}[b]{0.5\linewidth}

\centering
\resizebox{\linewidth}{!}{
\setlength{\tabcolsep}{2pt}
\begin{tabular}{cccccccc}
\small
\rotatebox[origin=l]{90}{\hspace{-0.2cm} 3 Dancers} &
\shortstack{\includegraphics[width=0.33\linewidth]{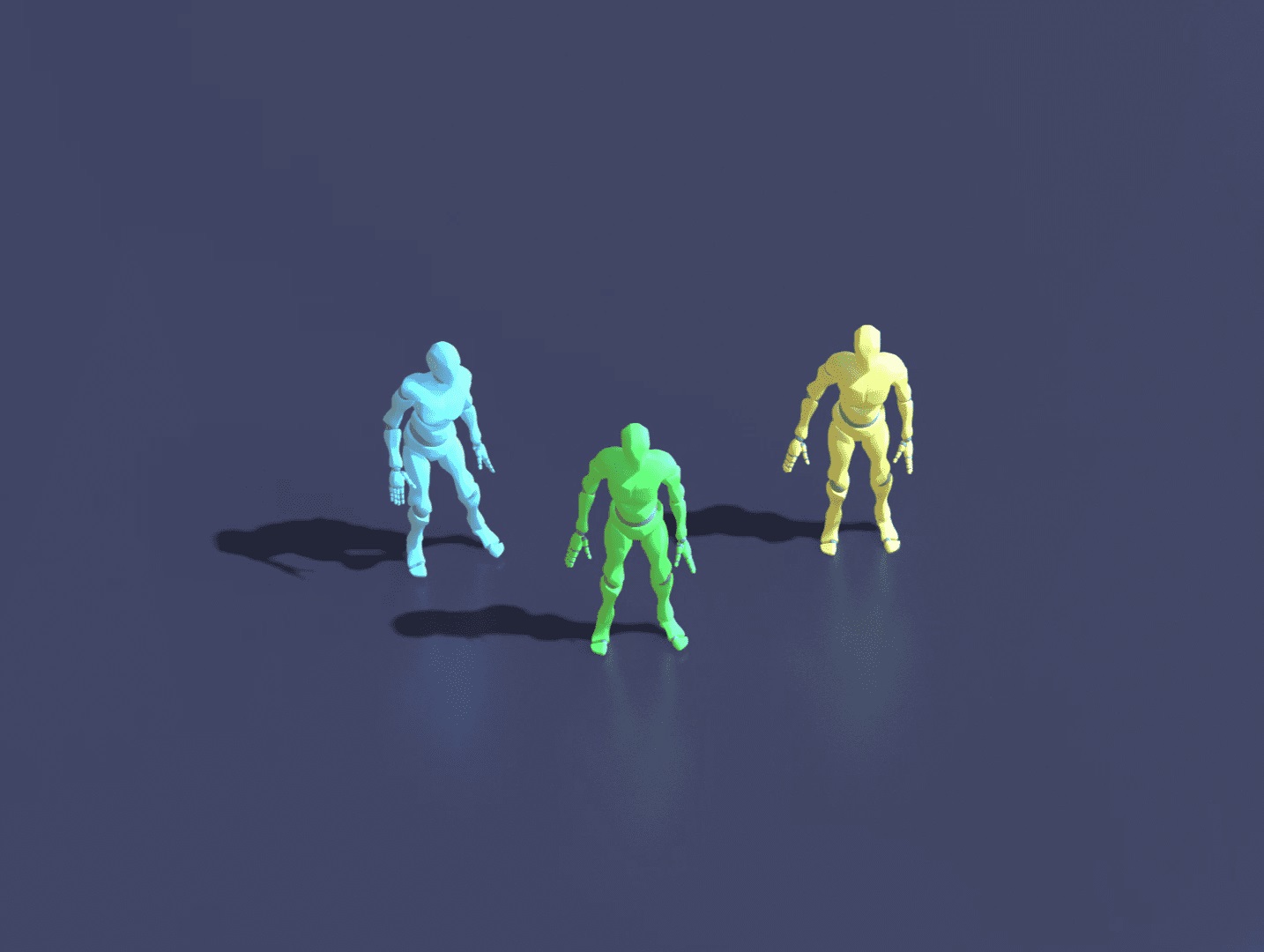}}&
\shortstack{\includegraphics[width=0.33\linewidth]{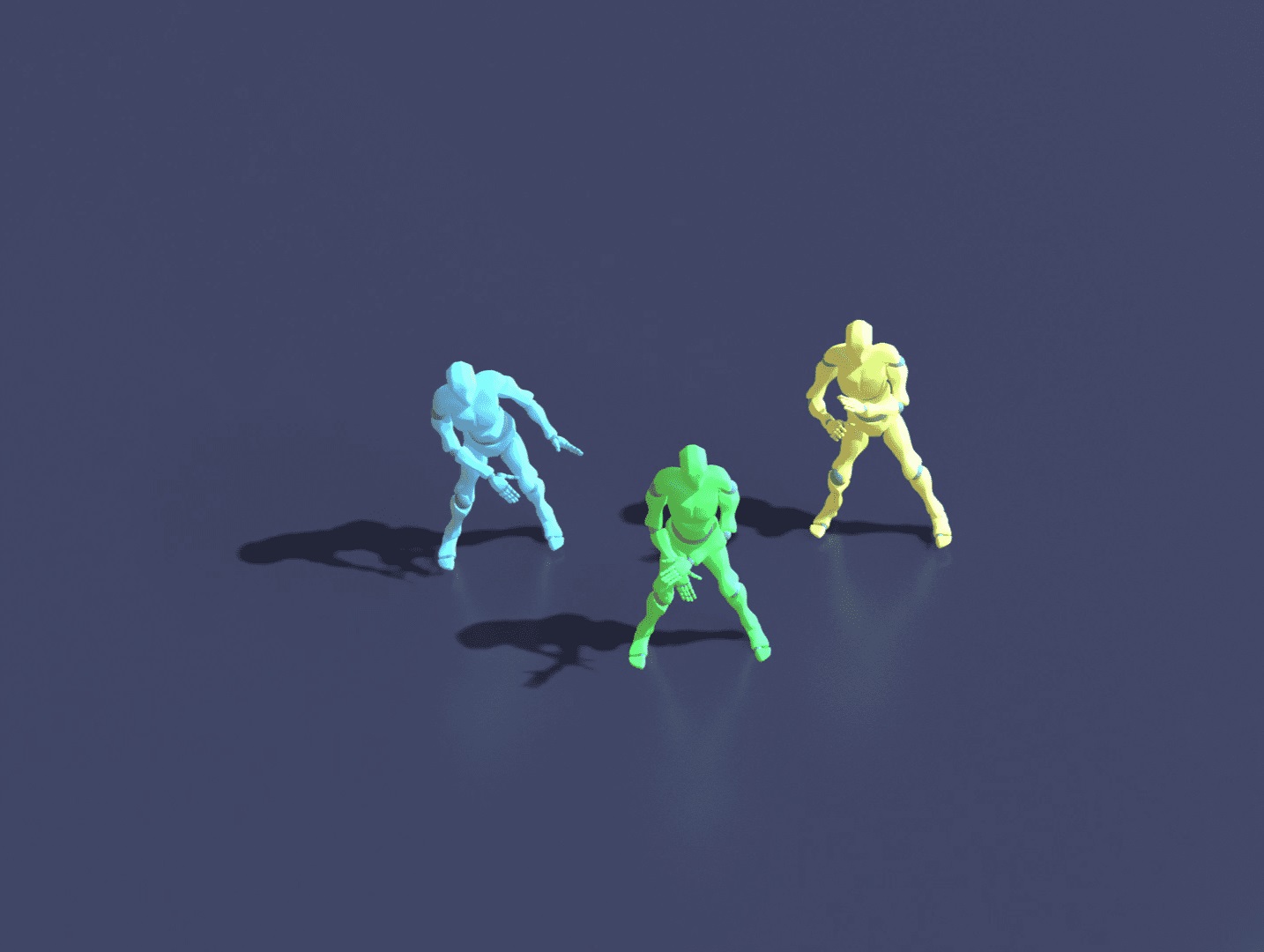}}&
\shortstack{\includegraphics[width=0.33\linewidth]{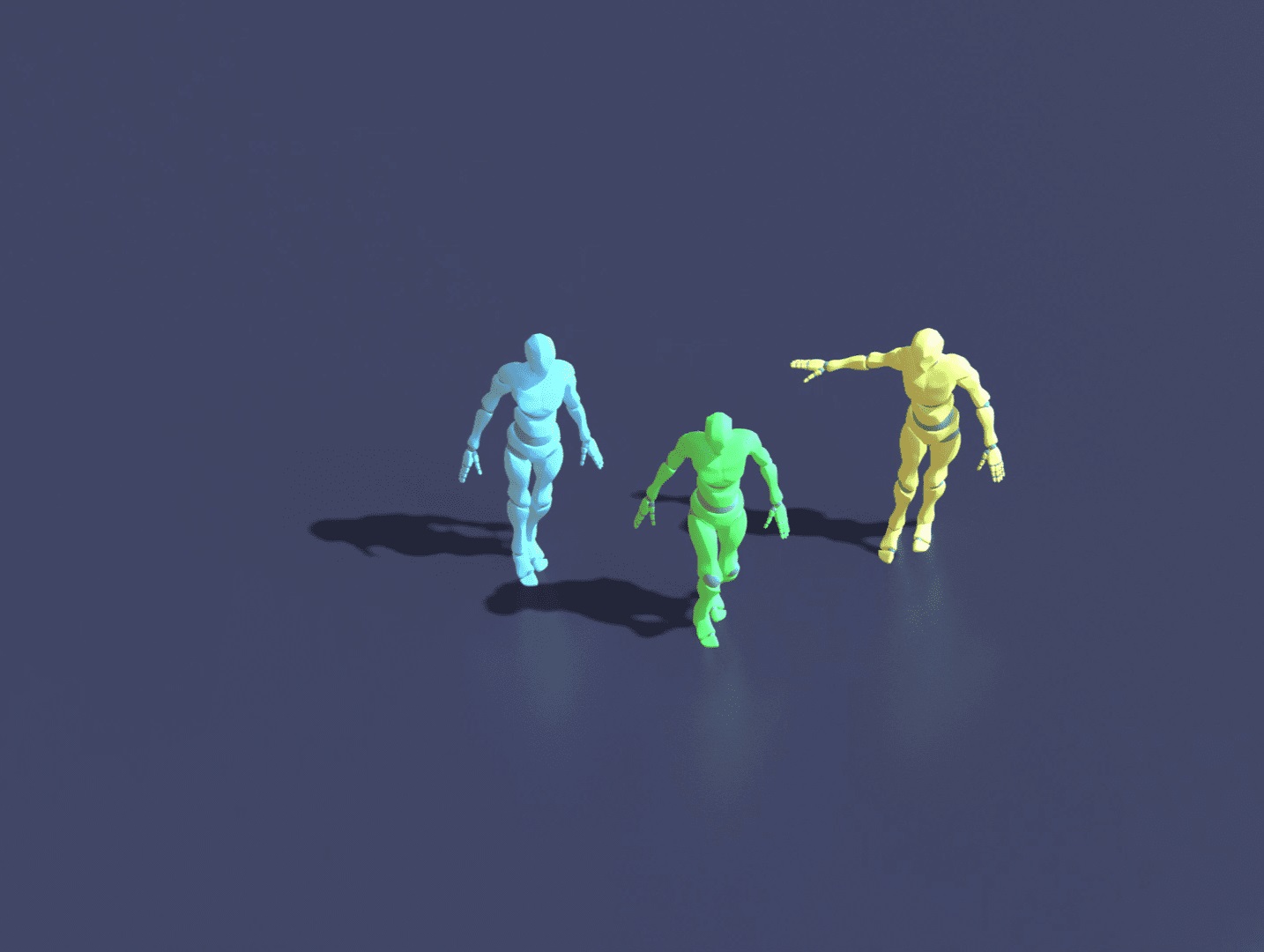}}\\[3pt]
\rotatebox[origin=l]{90}{\hspace{-0.2cm} 5 Dancers} &
\shortstack{\includegraphics[width=0.33\linewidth]{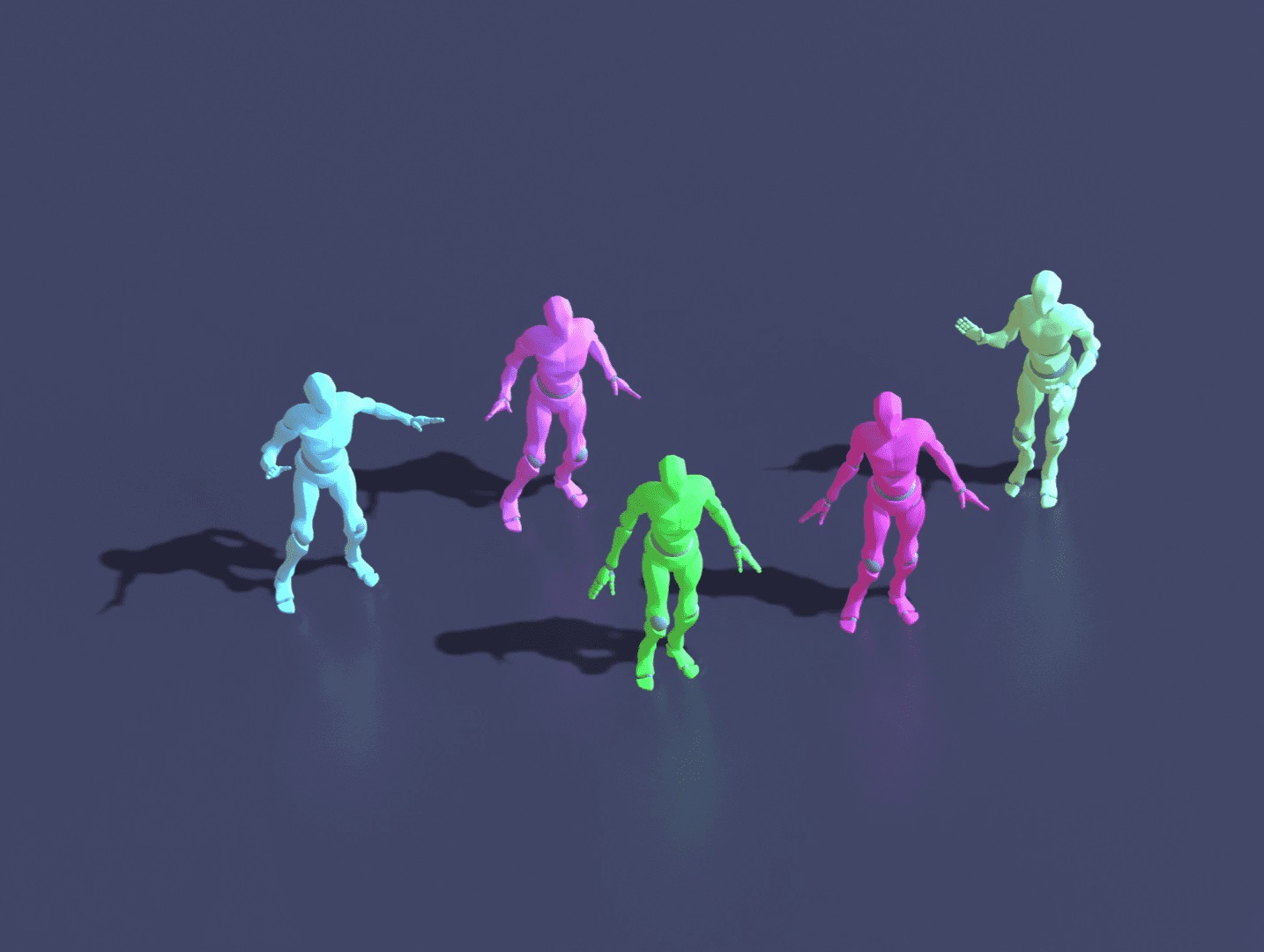}}&
\shortstack{\includegraphics[width=0.33\linewidth]{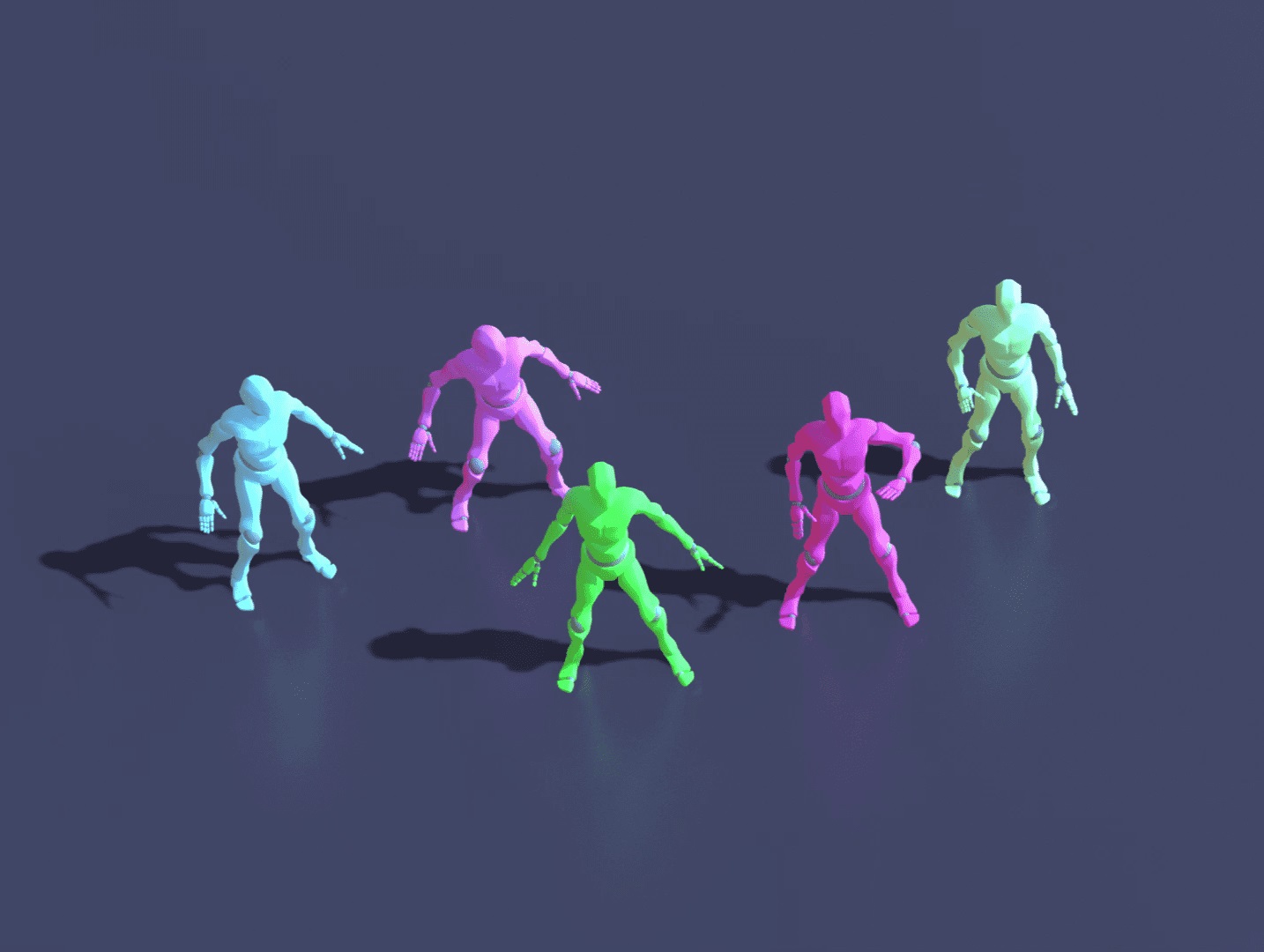}}&
\shortstack{\includegraphics[width=0.33\linewidth]{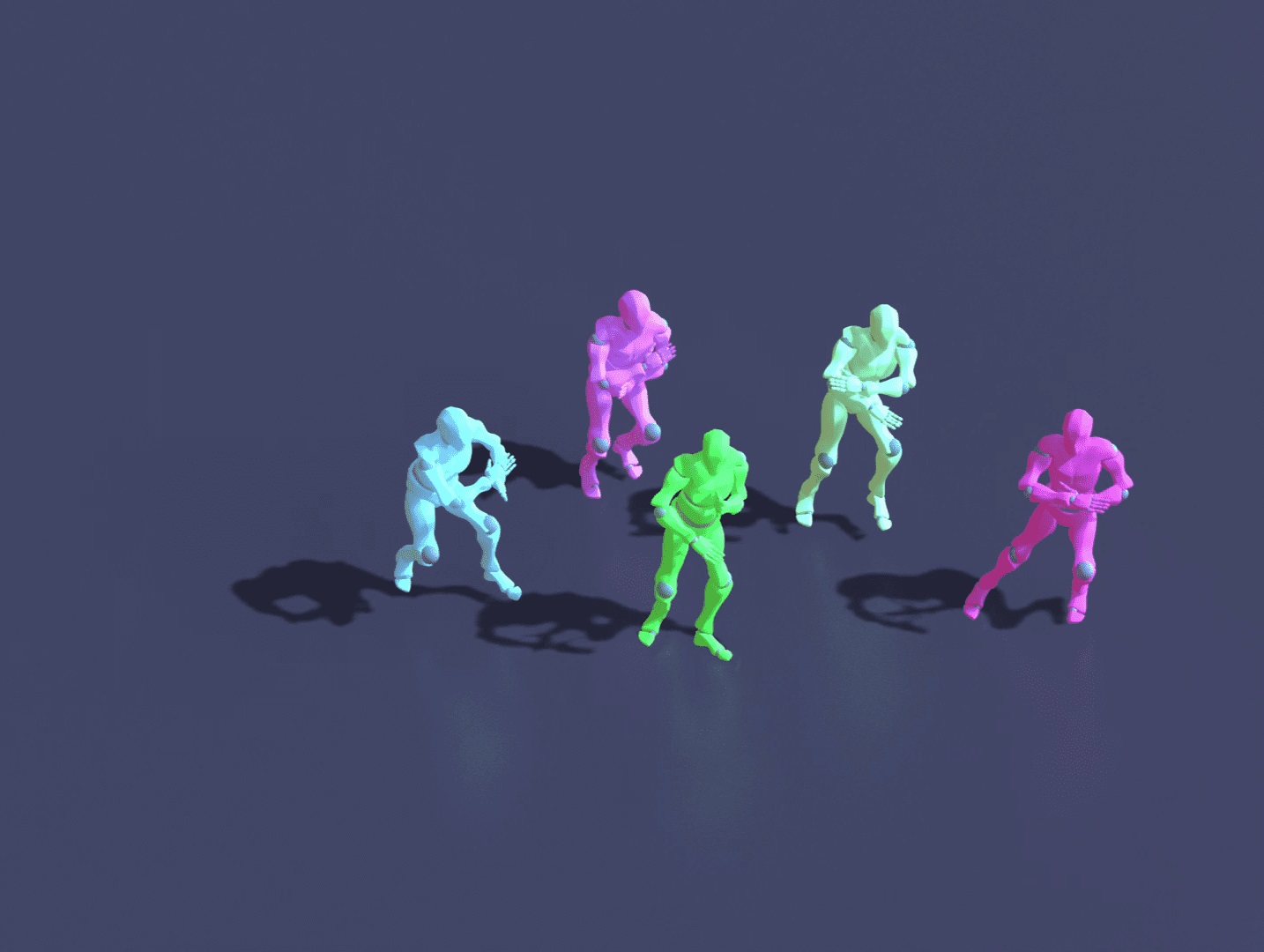}}\\[3pt]
\rotatebox[origin=l]{90}{\hspace{-0.2cm} 7 Dancers} &
\shortstack{\includegraphics[width=0.33\linewidth]{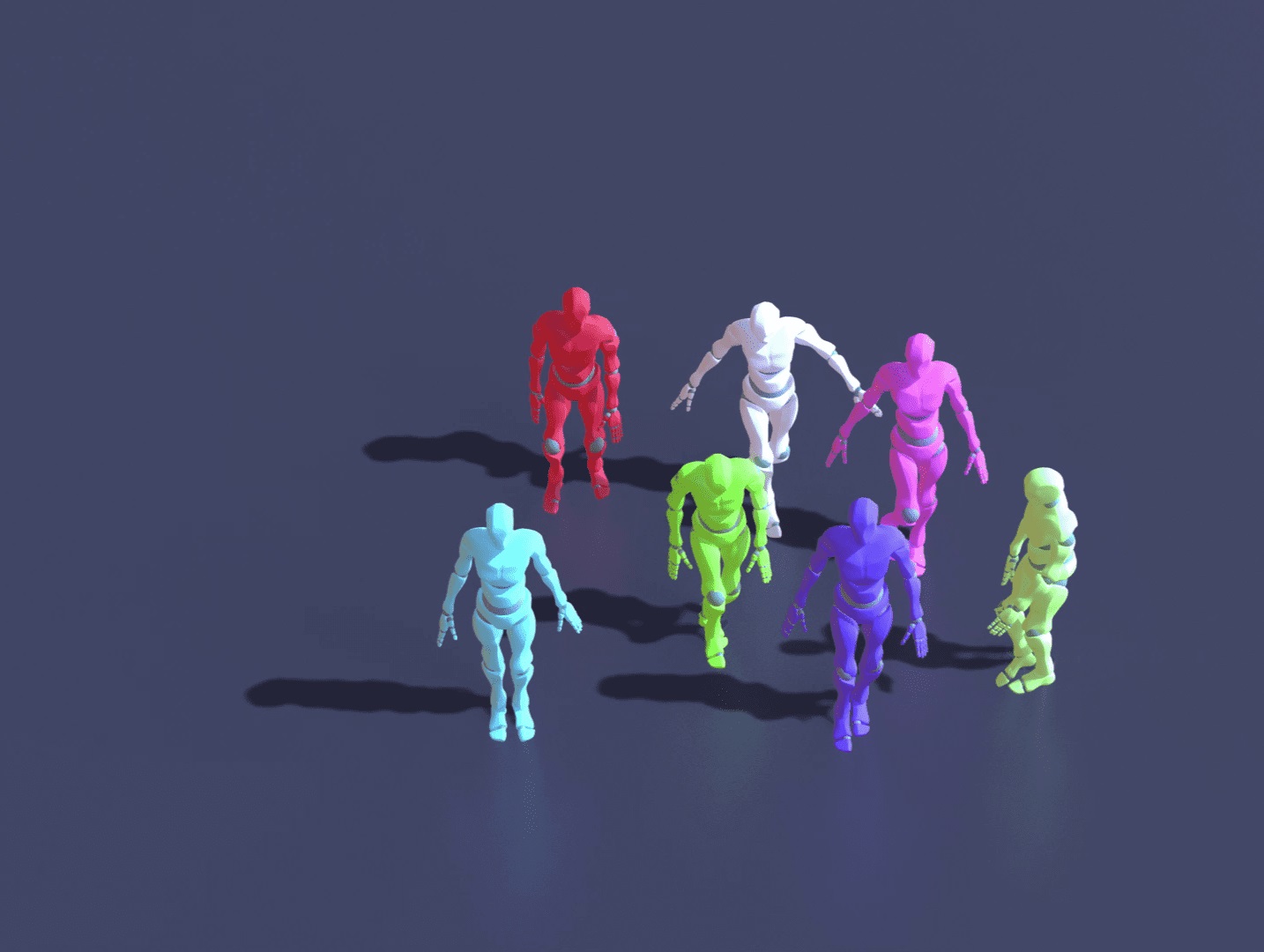}}&
\shortstack{\includegraphics[width=0.33\linewidth]{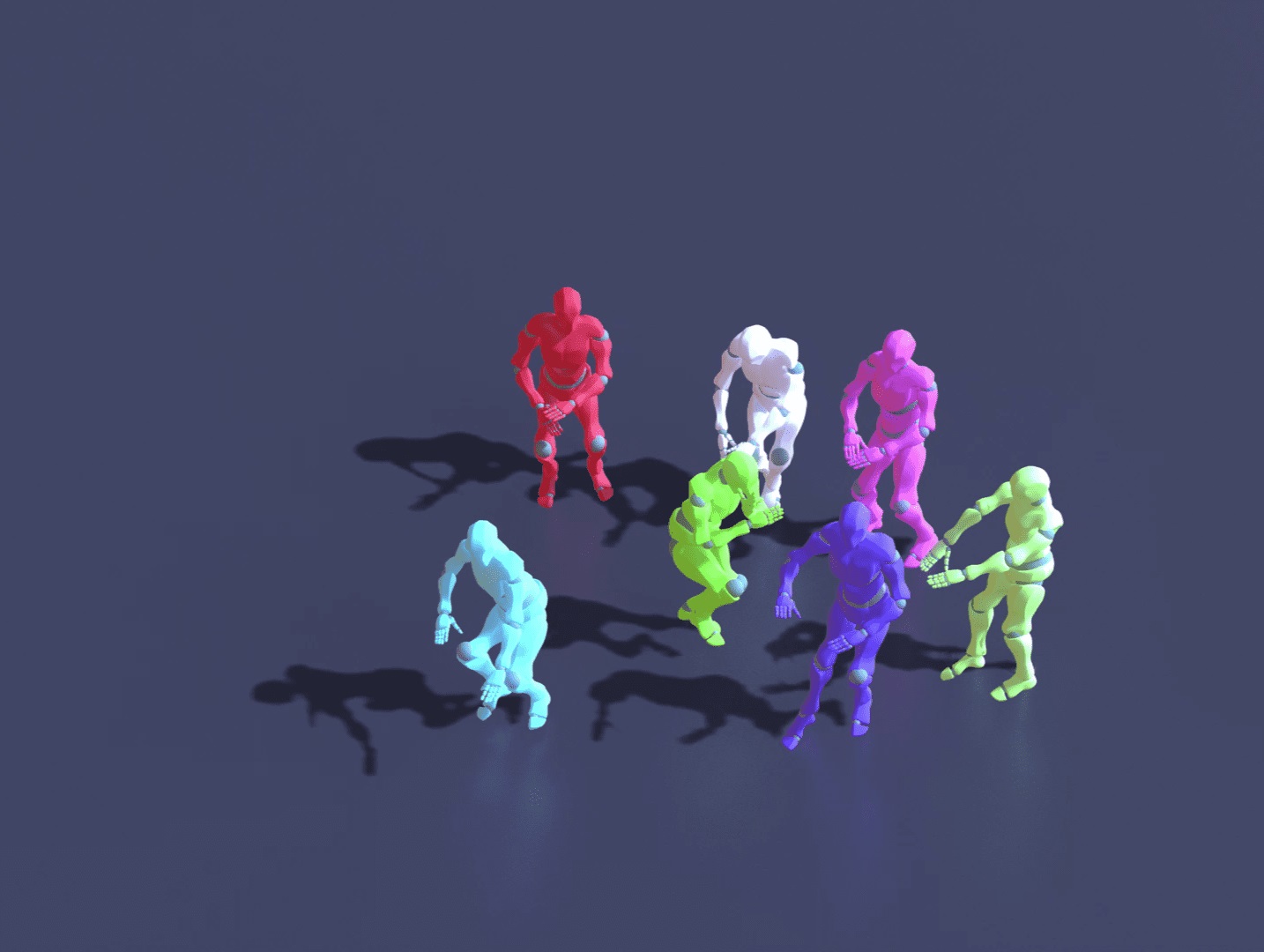}}&
\shortstack{\includegraphics[width=0.33\linewidth]{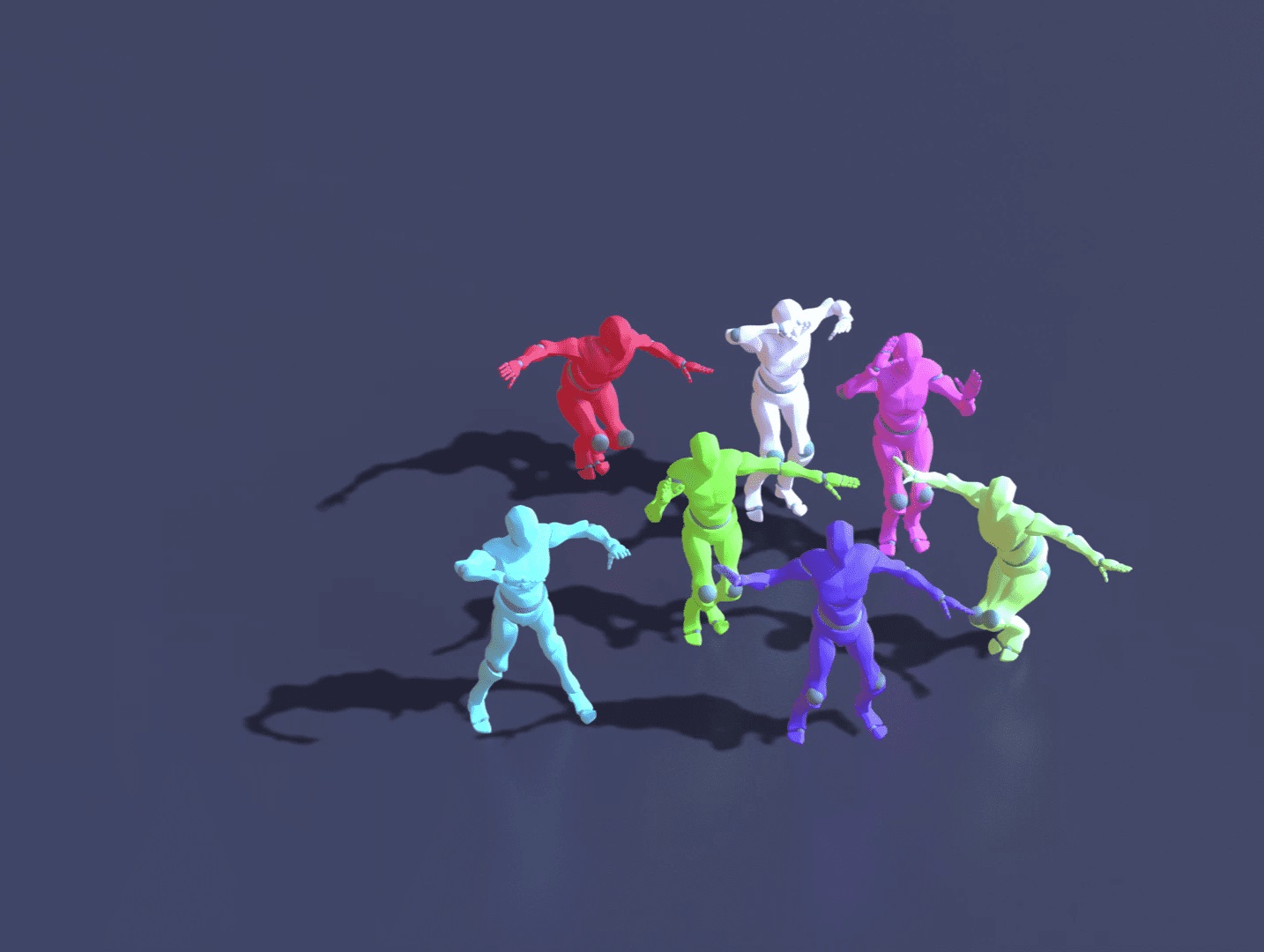}}\\[3pt]
\rotatebox[origin=l]{90}{\hspace{-0.3cm} 10 Dancers} &
\shortstack{\includegraphics[width=0.33\linewidth]{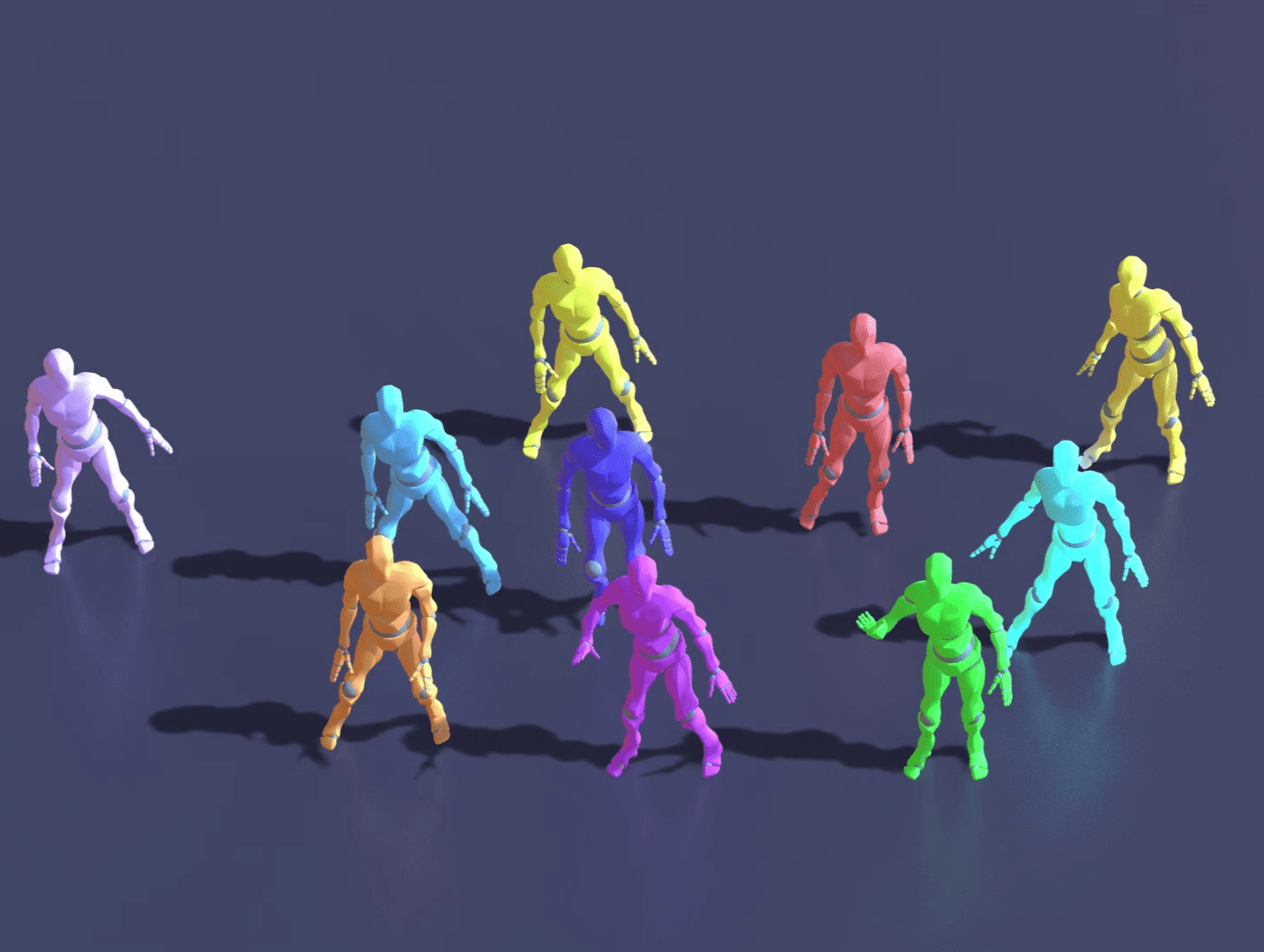}}&
\shortstack{\includegraphics[width=0.33\linewidth]{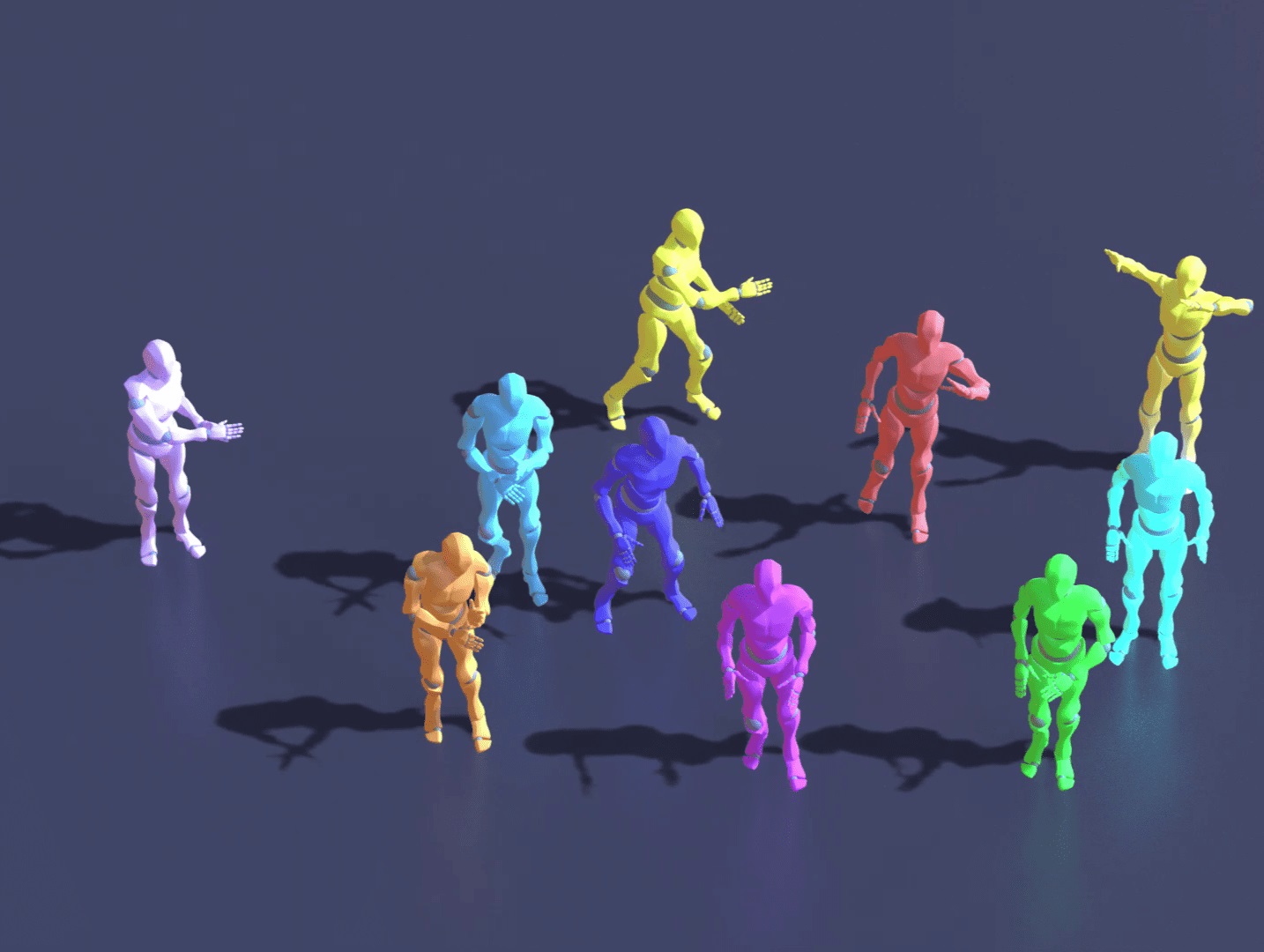}}&
\shortstack{\includegraphics[width=0.33\linewidth]{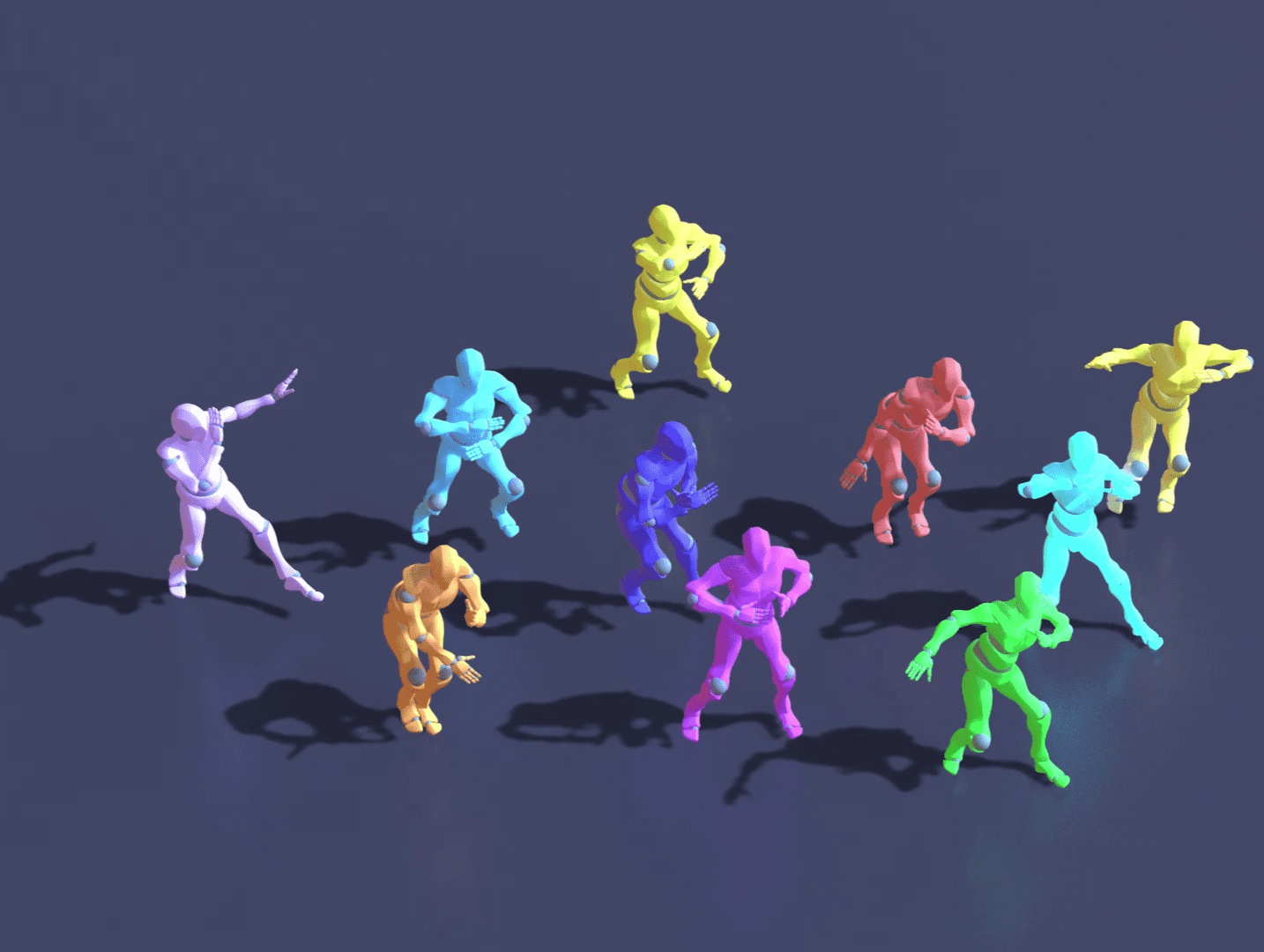}}\\[1pt]

\end{tabular}
}
\vspace{1ex}
    \captionof{figure}{Visualization of scalable dancers.}
    \label{fig:ScaleVis}

\end{minipage}
\hfill
\begin{minipage}[b]{0.47\linewidth}
\centering
\includegraphics[width=1\textwidth, keepaspectratio=true]{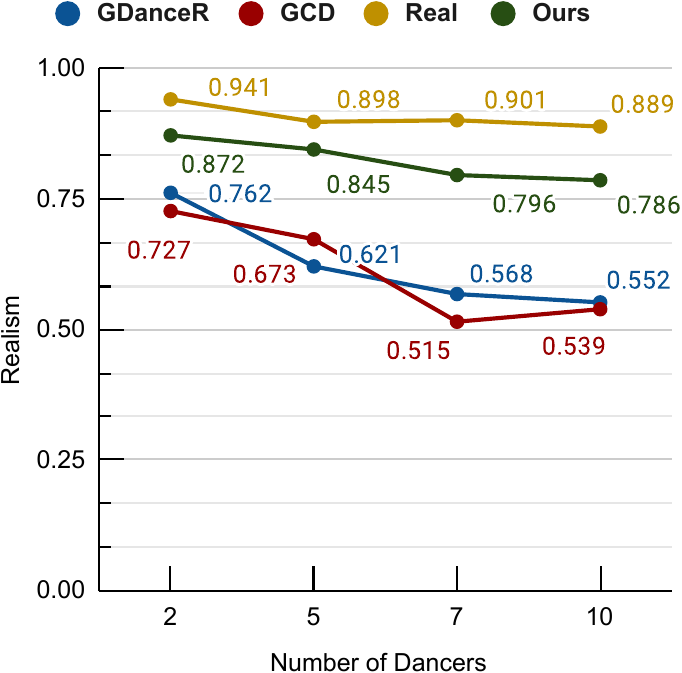}
\captionof{figure}{Realism between different methods when number of dancers is varied.}
\label{fig:UserStudy2}
\end{minipage}
\end{table}
\vspace{-10ex}

\section{Discussion}
While our approach leverages the VAE as a primary solution for generating a manifold, it is important to acknowledge certain inherent limitations associated with this choice. One notable challenge is the susceptibility to issues such as posterior collapse and unstable sampling within the VAE framework. These challenges can result in generated group dance motions that may not consistently meet performance expectations.

One specific manifestation of this limitation is the potential for false decoding when sampling points that lie too far from the learned distribution. This scenario can lead to unexpected rotations or disruptions in the physics of the generated content. The impact of this problem becomes evident in instances where the generated samples deviate significantly from the anticipated distribution, introducing inaccuracies and distortions.

To address these challenges, we recognize the need for ongoing efforts to mitigate the effects of posterior collapse and unstable sampling. While the problem is acknowledged, our approach incorporates measures to limit its impact. Future research directions could explore alternative generative models or additional techniques to enhance the robustness and reliability of the generated results in the face of these identified limitations. 

%
%
\bibliographystyle{splncs04}
\bibliography{acmart}
\end{document}